\documentclass[runningheads]{llncs}

\usepackage{eccv}

\usepackage{eccvabbrv}

\usepackage{graphicx}
\usepackage{booktabs}

\usepackage[accsupp]{axessibility}  

\usepackage{hyperref}

\usepackage{orcidlink}

\usepackage{multirow}
\usepackage{pifont}  
\usepackage{xcolor}   
\usepackage[utf8]{inputenc}
\usepackage{fontawesome}
\usepackage{makecell}

\usepackage{amssymb,bm}
\usepackage{algorithm}
\usepackage{algpseudocode}

\usepackage{amsmath}
\usepackage{colortbl}
\usepackage{pgfplots}
\pgfplotsset{compat=1.18}
\usepackage{pgfmath}
\usepackage{circledsteps}
\usepackage{comment}

\definecolor{myhigh}{HTML}{F1CCB5}
\definecolor{mylow}{HTML}{D4E2ED}

\newcommand{\colorize}[3]{
  \pgfmathsetmacro{\midpoint}{(#1 + #2)/2}%
  \pgfmathsetmacro{\scaled}{%
    ifthenelse(
      #3 < \midpoint,
      (#3 - #1)/(\midpoint - #1 + 0.0001), %
      (#3 - \midpoint)/(#2 - \midpoint + 0.0001) %
    )
  }%
  \pgfmathsetmacro{\colorvalue}{int(max(0, min(1, \scaled)) * 100)}%
  \pgfmathparse{#3 < \midpoint ? 1 : 0}%
  \ifnum\pgfmathresult=1
    \edef\temp{\noexpand\cellcolor{white!\colorvalue!mylow}}%
    \temp#3%
  \else
    \edef\temp{\noexpand\cellcolor{myhigh!\colorvalue!white}}%
    \temp#3%
  \fi
}

\newcommand{\colorizeinverse}[3]{%
  \pgfmathsetmacro{\midpoint}{(#1 + #2)/2}%
  \pgfmathsetmacro{\scaled}{%
    ifthenelse(
      #3 < \midpoint,
      (#3 - #1)/(\midpoint - #1 + 0.0001), %
      (#3 - \midpoint)/(#2 - \midpoint + 0.0001) %
    )
  }%
  \pgfmathsetmacro{\colorvalue}{int(max(0, min(1, \scaled)) * 100)}%
  \pgfmathparse{#3 < \midpoint ? 1 : 0}%
  \ifnum\pgfmathresult=1
    \edef\temp{\noexpand\cellcolor{white!\colorvalue!myhigh}}%
    \temp#3%
  \else
    \edef\temp{\noexpand\cellcolor{mylow!\colorvalue!white}}%
    \temp#3%
  \fi
}

\newcommand{\colorizeinverseLog}[3]{%

  \pgfmathsetmacro{\logmin}{ln(#1)}%
  \pgfmathsetmacro{\logmax}{ln(#2)}%
  \pgfmathsetmacro{\logval}{ln(#3)}%

  \pgfmathsetmacro{\midpoint}{(\logmin + \logmax)/2}%

  \pgfmathsetmacro{\scaled}{%
    ifthenelse(
      \logval < \midpoint,
      (\logval - \logmin)/(\midpoint - \logmin + 0.0001),
      (\logval - \midpoint)/(\logmax - \midpoint + 0.0001)
    )
  }%

  \pgfmathsetmacro{\colorvalue}{int(max(0, min(1, \scaled)) * 100)}%

  \pgfmathparse{\logval < \midpoint ? 1 : 0}%
  \ifnum\pgfmathresult=1
    \edef\temp{\noexpand\cellcolor{white!\colorvalue!myhigh}}%
    \temp#3%
  \else
    \edef\temp{\noexpand\cellcolor{mylow!\colorvalue!white}}%
    \temp#3%
  \fi
}

\newcommand{\corrauth}{\textsuperscript{\ensuremath{\dagger}}}

\begin{document}

\title{StreamEdit: Training-Free Video Editing via Few-Step Streaming Video Generation}

\titlerunning{StreamEdit}

\author{Guanlong Jiao\inst{1,4} \and 
Chenyangguang Zhang\inst{2} \and 
Jia Jun Cheng Xian\inst{1,4} \and \\
Zewei Zhang\inst{1,3} \and
Renjie Liao\inst{1,4,5} \corrauth}

\authorrunning{G.~Jiao, C.~Zhang, et al.}

\institute{
\footnotesize
$^{1}$ The University of British Columbia \quad
$^{2}$ ETH Zürich \quad
$^{3}$ McMaster University \\
$^{4}$ Vector Institute \quad
$^{5}$ Canada CIFAR AI Chair \quad \\
\email{
\{gjiao, anthony, rjliao\}@ece.ubc.ca, \\
chenyangguang.zhang@inf.ethz.ch, 
zhanz561@mcmaster.ca
}
}

\maketitle
\begingroup
\renewcommand{\thefootnote}{\ensuremath{\dagger}}
\footnotetext{Corresponding author.}
\endgroup

\begin{abstract}

Although existing video editing methods are generally feasible, they often require many costly iterations and still struggle to deliver high-quality yet satisfying editing results.
We attribute this limitation to the prevalent data-to-data paradigm, which is less compatible with modern generative models than noise-to-data generation.
To address this gap, we revisit video editing from a noise-to-data perspective and propose \textbf{Stream}ing-Generation-based Video \textbf{Edit}ing (\textbf{StreamEdit}), which preserves few-step sampling while seamlessly injecting source-video conditions.
Built on pre-trained streaming generation models, StreamEdit introduces dual-branch fast sampling with a self-attention bridge and cross-attention grounding/boosting to satisfy both sampling and conditioning requirements.
We further propose source-oriented guidance to improve target-generation quality, and a visual prompting strategy to enhance editing flexibility and practicality.
The method is effective, robust, and generalizable across different models.
Extensive experiments on diverse video editing tasks show that StreamEdit consistently outperforms existing approaches, even in few-step settings with minimal time cost.
Code and results are available at: \href{https://dsl-lab.github.io/StreamEdit/}{\texttt{dsl-lab.github.io/StreamEdit/}}.

  
\end{abstract}

\begin{figure*}[ht]
    \centering
    \includegraphics[width=0.99\linewidth]{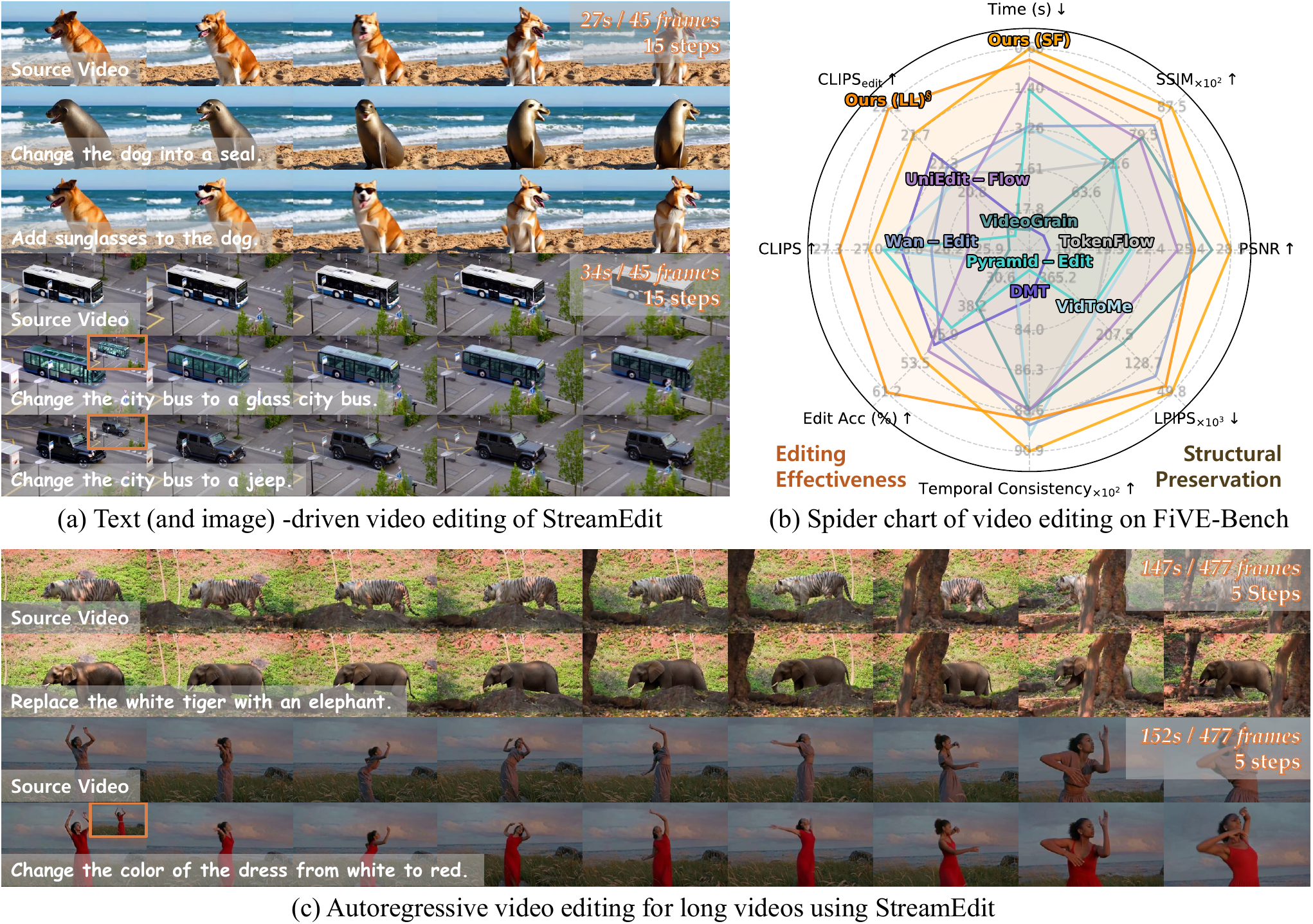}
    \caption{
    \textbf{Training-free text-driven (optional image-conditioned) streaming video editing of our proposed StreamEdit}.
    StreamEdit supports both text-driven video editing and first-frame-prompted editing, suitable for various editing tasks while having superiority in preservation of editing-irrelevant regions. 
    Developed on text-to-video streaming generation models, StreamEdit supports efficiently video editing for any length, with less than 0.32 second per frame (5 steps) on a single A100 GPU.
    }
    \label{fig: teaser}
\end{figure*}

\section{Introduction}

With the rapid progress of iterative visual generation models \cite{esser2024scaling, flux2024, wan2025, xiang2025structured}, training-free visual editing has achieved promising results in images \cite{sdedit, ju2023direct, kulikov2024flowedit, jiao2025uniedit}, 3D assets \cite{bar2025editp23, zhou2025anchorflow}, and videos \cite{qi2023fatezero, kara2024rave, liu2024video, wang2025videodirector}.
This progress has also enabled broader applications, including virtual reality \cite{song2025worldforge, jeong2025reangle}, embodied intelligence \cite{li2025mimicdreamer, dong2025emma}, \etc.

Mainstream training-free editing follows a \textbf{data $\to$ data} paradigm and is commonly divided into inversion-based \cite{masactrl, wang2024taming, jiao2025uniedit} and inversion-free methods \cite{infedit, kulikov2024flowedit, kim2025flowalign}.
Inversion reverses the generation process to recover reconstructive noise \cite{song2020denoising, mokady2023null, jiao2025uniedit}.
Accordingly, inversion-based methods perform \textbf{source $\to$ noise $\to$ target} transfer, but incur extra iterations and error accumulation from inversion, which often causes distortion in editing-irrelevant regions for complex videos (\eg, UniEdit-Flow in Fig. \ref{fig: Qualitative comparisons}).
Inversion-free methods instead perform direct \textbf{source $\to$ target} transfer with noisy auxiliaries \cite{kulikov2024flowedit, kim2025flowalign}.
However, because early samples remain close to the source, visible edits usually emerge only after many iterations.
Moreover, their irregular sampling trajectories typically require small, careful step sizes, leading to slow and unstable behavior on challenging cases \cite{kim2025flowalign}.
Overall, although these methods are feasible, video editing remains constrained by high computational cost and limited edit quality.

This limitation motivates a different view: video editing is fundamentally a video-conditioned video generation problem \cite{cheng2023consistent, vace, wu2025insvie}.
Modern generators follow the \textbf{noise $\to$ data} paradigm and are increasingly optimized for few-step, high-quality inference \cite{song2020denoising, liu2023instaflow, podell2023sdxl}.
Representative advances, including consistency models \cite{song2023consistency}, rectified flow \cite{liu2022flow}, and distribution matching distillation \cite{yin2024one}, have substantially accelerated generation.
Yet these benefits are rarely transferred to editing, because data-to-data pipelines are tightly coupled to their own iterative procedures and therefore cannot directly exploit fast few-step sampling \cite{song2023consistency}.

Our key insight is to reformulate editing as source-conditioned \textbf{noise $\to$ target} generation.
This reformulation preserves the few-step sampling while injecting source-video conditions, enabling faster and higher-quality editing.

To this end, we propose \textbf{Stream}ing-Generation-based Video \textbf{Edit}ing, termed \textbf{StreamEdit}.
As high-quality editing requires both target transformation and faithful preservation of structure, motion, and background, we build StreamEdit on pre-trained streaming generators with a dual-branch few-step sampler that produces parallel source and target features under the shared noise.
We then propose a self-attention bridge, combined with cross-attention grounding, to transfer source priors for structural and motion consistency and background preservation.
To improve controllability, we further design cross-attention regional boosting as a valve for editing strength.
To reduce stochastic instability, we add source-oriented guidance that better aligns editing-irrelevant regions with the source video.
Beyond text-only editing, we also provide an optional visual prompting mechanism using a target first frame to improve detail control and reliability.

Overall, StreamEdit establishes a training-free, generation-style paradigm for efficient and effective few-step streaming video editing.
Our contributions can be summarized as follows:
\begin{itemize}
    \item[$\bullet$] We reformulate training-free video editing as few-step streaming generation and introduce a corresponding sampling-and-guidance framework.
    \item[$\bullet$] We design dual-branch sampling with a self-attention bridge and cross-attention grounding/boosting to inject source conditions seamlessly.
    \item[$\bullet$] We introduce optional visual prompting, and finally enabling efficient, effective, flexible, and length-unrestricted streaming video editing.
    \item[$\bullet$] As shown in Fig. \ref{fig: teaser}, extensive experiments and user studies demonstrate consistent superiority of StreamEdit, supporting a viable new paradigm for training-free video editing.
\end{itemize}

\section{Related work}

\subsection{Image and Video Editing}
Modern generative models are predominantly built on diffusion \cite{ho2020denoising, song2020denoising} or flow matching \cite{lipman2022flow,liu2022flow,albergo2023stochastic}, which iteratively map noise distributions to data distributions.
Training-free visual editing typically follows a distinct iterative data $\to$ data paradigm \cite{sdedit, avrahami2022blended, prompt2prompt}, and mainly includes inversion-based and inversion-free approaches.
Inversion-based methods first add noise to the source sample and then denoise under target conditions, implicitly performing data transfer through source $\to$ noise $\to$ target \cite{mokady2023null, ju2023direct, dong2023prompt, geyer2024tokenflow, rout2024rfinversion, deng2024fireflowfastinversionrectified, wang2025videodirector, jiao2025uniedit}.
Inversion-free methods instead perform direct source $\to$ target editing, usually by leveraging noised samples to approximate transfer trajectories \cite{infedit, kulikov2024flowedit, kim2025flowalign, gao2025flowportal}.
Attention manipulation is also widely used to improve editing controllability and reliability \cite{masactrl, wang2024taming, ouyang2025proedit, bai2025uniedit, li2025flowdirector}.
Training-based editing has also advanced, but often depends on training-free methods for data preparation and inherits corresponding bottlenecks \cite{brooks2023instructpix2pix, Zhang2023MagicBrush, cheng2023consistent, wu2025insvie, zi2025senorita}.
Although training-free methods are applicable to both images and videos, the higher complexity and dimensionality of video data typically lead to slower inference and reduced editing quality.

\subsection{Streaming Video Generation}
Streaming generation refers to real-time, chunk-wise dynamic generation; in video generation, it is commonly realized through temporal autoregression \cite{chen2024diffusion, zhang2025packing, guo2025long, li2025stable}.
These methods condition each chunk on previously generated frames to produce temporally coherent long videos.
Recently, many works distill short-video generators into autoregressive models, preserving pre-trained generation quality and speed while introducing previous-frame conditioning with minimal architectural change \cite{xie2025progressive, cui2025self, yin2025slow}.
In addition, similar to key-value (KV) caching in language models \cite{vaswani2017attention, kwon2023efficient}, injecting previous conditions via visual KV modification in attention blocks has become a standard design choice \cite{huang2025self, yang2025longlive, ji2025memflow}.
These distribution-matching-distilled approaches generally support fast few-step video generation.
However, constrained by existing training-free editing paradigms, these advances have not yet translated into substantially faster and more effective video editing.

\section{Preliminary}

\subsection{Flow Matching}

Most of the existing high-quality generation models are based on diffusion \cite{ho2020denoising, song2020denoising, rombach2022high}, which learns the transformation from known distributions ($\pi_1$, \eg, $\mathcal{N}(0,I)$) to complex data distributions $\pi_0$.
One special case, flow matching \cite{lipman2022flow, liu2022flow, albergo2023stochastic}, has gained significant traction and is widely applied to video generation tasks \cite{jin2024pyramidal, wan2025, polyak2024movie} due to its effectiveness and simple training objective:
\begin{equation}
    \min_{\theta}~ \mathbb{E}_{x_0, x_1, t} \left[ \left\Vert \left( x_1 - x_0 \right) - {v}_{\theta} \left( x_t, t \right) \right\Vert^2 \right], \quad
    x_t = (1-t) x_0 + tx_1, \quad t\in[0, 1],
\end{equation}
which aims to learn a straight trajectory between data $x_0 \in \pi_0$ and noise $x_1 \in \pi_1$ by training the velocity function $v_\theta$ with $x_1 - x_0$ as the objective.

\subsection{Consistency Models}

Consistency models are designed for fast generation, establishing the direct mapping between noise and data \cite{song2023consistency}.
The sampling strategy of consistency models is widely adopted by few-step models \cite{luo2023latent, sabour2025align}.
It first predicts the clean data from a noisy sample, then obtains the next-step noisy data via a linear interpolation from the prediction and a random noise. 
Its flow-based formula is:
\begin{equation}
    z_{t_{i-1}} = x_{t_{i}} - t_{i}v_\theta(x_{t_{i}}, t_{i}), \quad
    x_{t_{i-1}} = (1-t) z_{t_{i-1}} + t_{i-1} \epsilon, \quad 
    \epsilon \sim \mathcal{N}(0, 1),
\label{eq:consistency_model}
\end{equation}
where $z_{t_{i-1}}$ is the one-step prediction of clean data from the previous noisy sample $x_{t_{i}}$, the final output is $z_{t_0}$, and $i \in [N]$, $t_N=1$, $t_0=0$, $x_{t_N} \sim \mathcal{N}(0, 1)$.

\subsection{KV Caching in Streaming Video Generation}

To reduce redundant computations, recent work has increasingly favored storing and incorporating prior information as a KV cache through one-time computation \cite{huang2025self, yang2025longlive, ji2025memflow}, as opposed to methods using overlapping sliding windows \cite{chen2024diffusion, zhang2025videomerge, cai2025ditctrl}.
Take Self Forcing \cite{huang2025self} and LongLive \cite{yang2025longlive} as examples, these methods conduct one model forward step using previously generated clean data as the input, caching its KV for providing previous condition and then injecting previous KV into self-attention, thus making the streaming generation coherent and consistent.

\section{Method}

\begin{figure*}[t]
    \centering
    \includegraphics[width=0.99\linewidth]{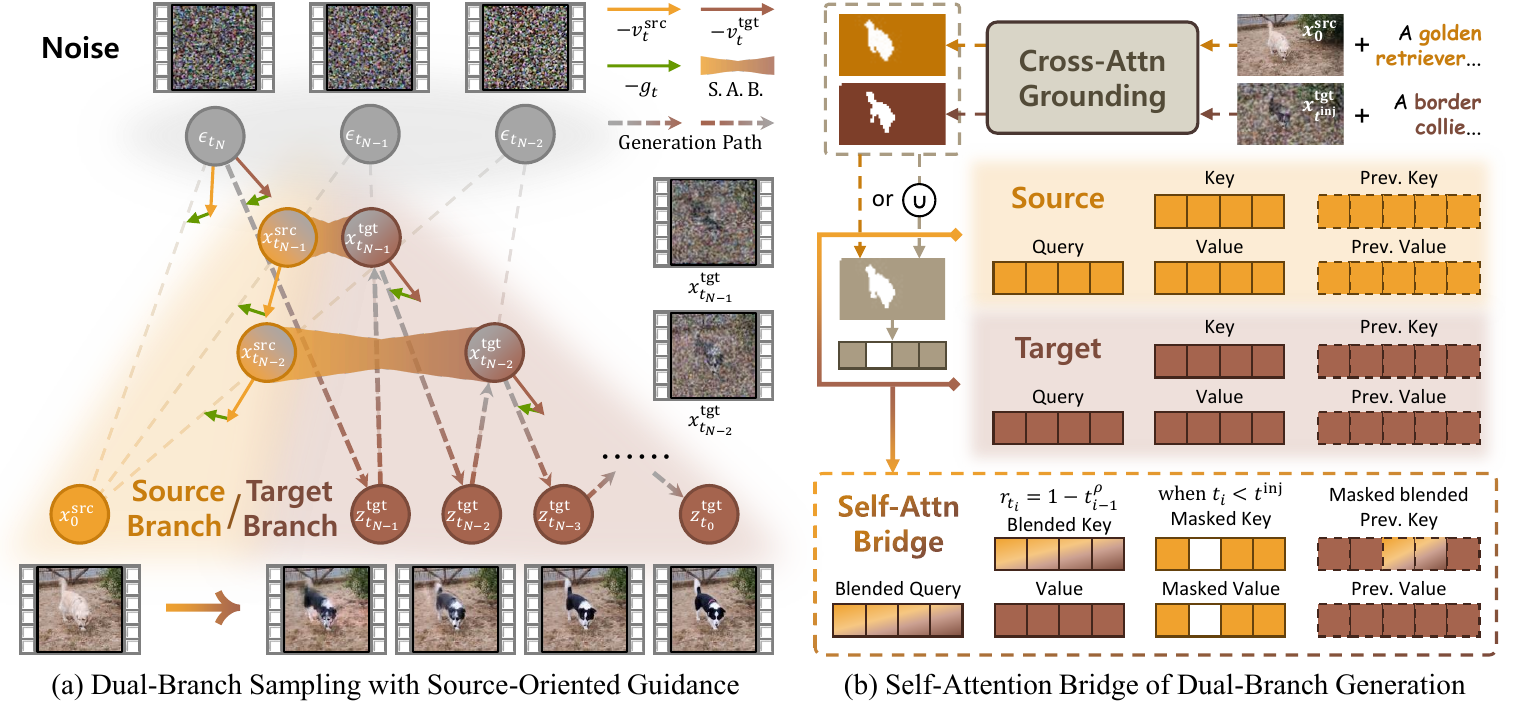}
    \caption{
    \textbf{Overview} of the framework and components of our proposed StreamEdit.
    $z^\text{tgt}_{t}$ means the one-step prediction at timestep $t$. 
    $x^\text{src}_t$ indicates the linear interpolation of $x_0^\text{src}$ and $\epsilon_t$ while $x^\text{tgt}_t$ denotes it of $z_{t}^\text{tgt}$ and $\epsilon_t$.
    (a) illustrates the generation process and its corresponding samples.
    We use different colored regions to distinguish the dual branches.
    Dashed arrow lines depict the stochastic few-step sampling process of target generation.
    The green arrow $-g_t$ indicates source-oriented guidance, where the latent mask is omitted here.
    (b) presents self-attention bridge (S.A.B.) for source condition injection and the mask obtained from cross-attention grounding as editing-related regions.
    The trigger words used for grounding are manually provided by users.
    }
    \label{fig: framework}
\end{figure*}

We reformulate video editing as fast streaming generation.
Built on representative streaming generators, including Self Forcing \cite{huang2025self} and LongLive \cite{yang2025longlive}, our method develops source-video-conditioned target generation through few-step sampling and attention manipulation.
First, we extend stochastic few-step sampling to a dual-branch framework that propagates noised source information in parallel with the target sample (Fig. \ref{fig: framework} (a)).
We then design a self-attention bridge to inject source conditions into target generation through the model's dynamic attention mechanism (Fig. \ref{fig: framework} (b)).
To suppress stochastic artifacts in editing-irrelevant regions, we introduce source-oriented guidance (green arrow in Fig. \ref{fig: framework} (a)), analogous to classifier-free guidance for quality improvement.
We further introduce cross-attention grounding and boosting, which provide the mask for the self-attention bridge and a controllable valve for editing strength.
In addition, we propose visual prompting to enhance fine-grained visual control.
Together, these components form a streaming-generation-style pipeline for fast, controllable, and effective video editing.

Our method has two core components: sampling strategy (Sec. \ref{sec:Dual-Branch Fast Sampling} and \ref{sec:Source-Oriented Guidance}) and attention manipulation (Sec. \ref{sec: self-attention bridge} and \ref{sec: cross-attn}).
Sec. \ref{sec:Dual-Branch Fast Sampling} presents the sampling paradigm.
Sec. \ref{sec: self-attention bridge} and \ref{sec: cross-attn} describe the attention designs.
Sec. \ref{sec:Source-Oriented Guidance} introduces sampling guidance, and Sec. \ref{sec:Visual Prompting} presents visual prompting for practical control.

\subsection{Dual-Branch Fast Sampling}
\label{sec:Dual-Branch Fast Sampling}

Inversion-based editing methods typically incur high computational cost and cumulative errors due to imperfect inversion, often resulting in poor background preservation \cite{ju2023direct, mokady2023null, garibi2024renoise, wang2024taming}.
Inversion-free methods, in contrast, progressively edit from the source video over many iterations, but often remain limited in both efficiency and effectiveness \cite{infedit, kulikov2024flowedit, kim2025flowalign}.
To address these limitations, we reformulate video editing as conditional generation through training-free operations, leveraging the few-step capability of distilled models \cite{song2023consistency, yin2024one, sauer2024adversarial}.

To this end, we first extract source-video representations that can be propagated in parallel with target generation during few-step sampling \cite{song2023consistency}.
Specifically, we treat intermediate states from the source sampling process as features that can be synchronously injected into target generation, yielding a dual-branch framework.
Given a source video with its source prompt and target prompt, we denoise source and target samples in parallel to provide source information for editing.
Under stochastic few-step sampling, we formulate the source branch ($\text{src}$) as fixed-endpoint sampling with a constant clean sample:
\begin{equation}
\begin{split}
    &x_{t_i}^\text{src}=(1-t_i)x_0^\text{src}+t_i\epsilon_{t_i}, \quad
     x^\text{tgt}_{t_i}=(1-t_i)z_{t_i}^\text{tgt} + t_i\epsilon_{t_i}, \quad 
     \epsilon_{t_i} \sim \mathcal{N}(0, 1),  \\
    &v_{t_i}^\text{src}=v_\theta(x_{t_i}^\text{src},t_i), \quad 
     v_{t_i}^\text{tgt}=v^*_\theta(x_{t_i}^\text{tgt},t_i ~\vert~ x_{t_i}^\text{src}), \quad
     i \in [1, N],
\end{split}
\label{eq:dual-branch}
\end{equation}
where $v^*_\theta$ is the target velocity ($\text{tgt}$) with the subsequent self-attention bridge and cross-attention boosting.
This framework satisfies the basic requirement for injecting source conditions into target generation at every denoising step.

\subsection{Self-Attention Bridge of Dual-Branch Generation}
\label{sec: self-attention bridge}

Self-attention is central to adaptive visual representation in DiT models \cite{dosovitskiy2020image, peebles2023scalable} and has been widely adopted in editing methods \cite{masactrl, infedit, yin2025consistedit}.
We therefore build a self-attention bridge between the source and target branches, enabling source-feature-conditioned target generation.
Given paired source-target attention features from dual-branch sampling, we design the bridge in three parts: query blending for structure and motion, key blending for editing effectiveness and consistency, and source KV injection for background preservation (Fig. \ref{fig: framework} (b)).

\begin{figure*}[t]
    \centering
    \includegraphics[width=0.99\linewidth]{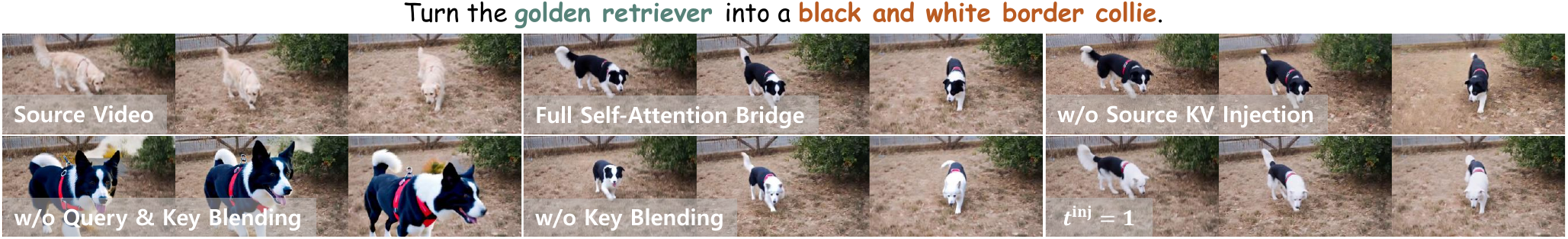}
    \caption{
    \textbf{Visualization of the impact} of self-attention bridge's components.
    Query blending ensures basic structural preservation, while key blending keeps continuous editing effects.
    Source KV injection provides meticulous details for editing-irrelevant regions, with its delay mechanism ($t^\text{inj}=1$) preventing editing failures.
    }
    \label{fig: vis self-attn bridge}
\end{figure*}

\subsubsection{Query Blending: Structure and Motion Preservation.}
Many attribute-editing tasks require changing object attributes while preserving global motion and trajectory.
To enforce this property, we blend source and target self-attention queries, since queries encode spatiotemporal structure and motion cues \cite{qi2023fatezero, tu2024motioneditor}.
Because early denoising steps mainly shape semantics and later steps refine details \cite{luo2023diffusion, tinaz2025emergence, yang2026stable}, we use a time-dependent blending ratio:
\begin{equation}
    \tilde{Q}^\text{tgt}_\text{SA} = r_{t_i} Q^\text{tgt}_\text{SA} + (1 - r_{t_i}) Q^\text{src}_\text{SA}, \quad
    r_{t_i} = 1 - t_{i-1}^{\rho},
\end{equation}
where $Q_\text{SA}$ denotes the original self-attention query across layers, $r_{t_i}$ is the blending ratio, and $\rho$ controls the preservation strength.
As $r_{t_i}$ increases from 0 to 1, source priors dominate early for structural guidance while later steps retain flexibility for target adaptation.

\subsubsection{Key Blending: Editing Effectiveness and Consistency.}
Query blending alone can still yield unstable edits when query-key matching is weak.
To stabilize attention and improve editing consistency, we propose to blend keys jointly with queries.
Meanwhile, to together strengthen the matching between queries and previous keys, we further apply similar blending to previous-frame keys.
Nevertheless, since source and generated target backgrounds are naturally similar in our framework, we propose to apply this alignment mainly to foreground regions to improve temporal consistency, and thus respectively design our key blending of current keys and masked blending of previous keys:
\begin{equation}
\begin{split}
    &\tilde{K}^\text{tgt}_\text{SA} = r_{t_i} K^\text{tgt}_\text{SA} + (1 - r_{t_i}) K^\text{src}_\text{SA}, \\
    &\tilde{K}^\text{tgt}_\text{prev} = M_\text{prev} \odot (r_{t_i} K^\text{tgt}_\text{prev} + (1 - r_{t_i}) K^\text{src}_\text{prev}) + (1-M_\text{prev}) \odot K^\text{tgt}_\text{prev},
\end{split}
\end{equation}
where $M_\text{prev}$ indicates the union mask of previous video chunks (introduced in Sec. \ref{sec: cross-attn}).
As shown in Fig. \ref{fig: vis self-attn bridge}, this design encourages edited regions to aggregate relevant information while reducing interference from unrelated regions, dropping it will lead to editing degradation (w/o Key Blending).

\subsubsection{Source KV Injection: Background Preservation.}
Despite query/key blending, precise background details are still insufficient without explicit source KV injection (Fig. \ref{fig: vis self-attn bridge}, w/o Source KV Injection).
However, injecting source KV throughout all timesteps can suppress target concept emergence (Fig. \ref{fig: vis self-attn bridge}, $t^\text{inj}=1$).
Empirically, source KV is most beneficial in low-noise steps, where denoising primarily refines details \cite{luo2023diffusion, tinaz2025emergence, yang2026stable}.
The final target-branch self-attention uses:
\begin{equation}
\begin{split}
    Q=\tilde{Q}_\text{SA}^\text{tgt}, \quad
    &K=[\tilde{K}^\text{tgt}_\text{SA}, \tilde{K}^\text{tgt}_\text{prev}, [t<t^\text{inj}]\cdot(M_\text{curr} \odot K^\text{src}_\text{SA})], \\
    &V=[V^\text{tgt}_\text{SA}, ~V^\text{tgt}_\text{prev}, ~[t<t^\text{inj}]\cdot(M_\text{curr} \odot V^\text{src}_\text{SA})],
\end{split}
\end{equation}
where $[t<t^\text{inj}]$ is the Iverson bracket; if the condition is false, the corresponding term is excluded from concatenation.
$t^\text{inj}$ denotes the timestep at which source KV injection starts and is fixed to 0.5 in our experiments.
$M_\text{curr}$ denotes the current-chunk mask (introduced in Sec. \ref{sec: cross-attn}).
Overall, the full self-attention bridge jointly preserves background and strengthens edit reliability (Fig. \ref{fig: vis self-attn bridge}, Full Self-Attention Bridge).

\subsection{Cross-Attention Grounding and Boosting}
\label{sec: cross-attn}

\subsubsection{Grounding: Editing-Related Mask.}
The self-attention bridge requires reliable editing-region masks, for which we design a cross-attention grounding module without introducing extra models.
Given user provided trigger words, prior works typically binarize trigger-word attention maps using fixed thresholds \cite{masactrl, yin2025consistedit}, which often require per-model or per-sample tuning.
Instead, we use foreground-background attention differences for adaptive grounding.
Given the cross-attention vector $A_\text{CA}(p,Q_\text{CA})$ for word $p$ in prompt $\mathcal{P}$ and query sequence $Q_\text{CA}$, we compute the trigger-word mask $\mathcal{T}$ from the difference between averaged attentions over word groups:
\begin{equation}
    M^{\phi} = H\left( 
    \frac{\sum_{p\in \mathcal{T}} A_\text{CA}^{\phi}(p,Q_\text{CA}^{\phi})}{\vert \mathcal{T} \vert} - 
    \frac{\sum_{p\in \mathcal{P} \setminus \mathcal{T}} A_\text{CA}^{\phi}(p,Q_\text{CA}^{\phi})}{\vert \mathcal{P} \setminus \mathcal{T} \vert}
    \right),
\end{equation}
where $\phi \in \{ \text{src}, \text{tgt} \}$ denotes the branch, $H(\cdot)$ is the Heaviside step function, and $\vert \cdot \vert$ denotes set cardinality.
We obtain $M^\text{src}$ at $t=0$ from the clean source latent, leveraging the distilled model's ability to retain spatial structure across timesteps.
For the target branch, $M^\text{tgt}$ is extracted at $t=t^\text{inj}$, after target semantics have formed.
We then compute the union mask
$M = \cup_{\phi} M^{\phi}$.
The union mask covers both source regions to be modified and target regions requiring editing.
Additionally, we denote the current-chunk mask as $M_\text{curr}$ and the masks from previous chunks as $M_\text{prev}$.

\subsubsection{Boosting: Regional Editing Enhancement.}
To provide a controllable valve for editing strength, we design cross-attention boosting based on the role of cross-attention in controlling prompt-word expression \cite{feng2022training, chefer2023attend}.
Multiplicative score enhancement can over-polarize token expression \cite{prompt2prompt, yin2024coloredit}; moreover, negative scores may unintentionally suppress trigger concepts.
Using masks from grounding, we therefore apply additive regional enhancement to trigger words inside editing regions, promoting edits while protecting background regions:
\begin{equation}
\begin{split}
    &A_\text{CA}^\text{tgt}(p,q)=\frac{\exp\left(S(p,q) + \ln w_{p,q}\right)}{\sum_{j\in Q_\text{CA}^\text{tgt}} \exp\left(S(j,q) + \ln w_{j,q}\right)}, \\
    &w_{p,q} = \begin{cases} 
        \omega, \quad \text{if } p \in \mathcal{T} \text{ and } M(q)=1 \\
        1, \quad \text{otherwise} 
        \end{cases},
\end{split}
\end{equation}
where $S(p,q)$ is the pre-softmax attention score between text token $p$ and visual query token $q$, and $\omega$ controls editing strength.
$M(q) \in \{0,1\}$ denotes the mask value of query token $q$; we use $M^\text{src}_\text{curr}$ in high-noise steps and $M_\text{curr}$ when $t<t^\text{inj}$.
This regional boosting improves controllability of edit intensity while reducing unintended background changes.

\subsection{Source-Oriented Guidance for Visual Quality}
\label{sec:Source-Oriented Guidance}
The self-attention bridge provides stable source-condition injection, while stochastic sampling improves editing flexibility and feasibility \cite{rout2024rfinversion, singh2024stochastic}.
However, stochasticity also makes the source branch a mixed trajectory of source video and random noise \cite{liu2022flow, lipman2022flow}, which introduces noisy source conditions and causes undesirable jitter in editing-irrelevant target regions, even with the self-attention bridge (Fig. \ref{fig: ablation studies} (b)).
This error is observable through the difference between the predicted source velocity and the ground-truth velocity of the corresponding linear interpolation:
\begin{equation}
    v^\text{gt}_{t_i} = \epsilon_{t_i} - x_0^\text{src}, \quad 
    g_{t_i} = v^\text{gt}_{t_i} - v_{t_i}^\text{src},
\label{eq:correction}
\end{equation}
where $g_{t_i}$ denotes the observable error.
Since randomness in editing-related regions remains beneficial for effective edits, we mainly apply correction to editing-irrelevant regions via a mask coefficient:
\begin{equation}
    \tilde{v}_{t_i}^\text{tgt} = v_{t_i}^\text{tgt} + \text{AMN}(v_{t_i}^\text{tgt} - v_{t_i}^\text{src}) \odot  g_{t_i}, \quad
    z^\text{tgt}_{t_{i-1}} = x^\text{tgt}_{t_i} - t_i \tilde{v}_{t_i}^\text{tgt},
\label{eq:velocity correction}
\end{equation}
where $\odot$ is the element-wise product and $\text{AMN}(\cdot)$ denotes the Abs-Mean-Norm operation (get the channel-wise mean of the input's absolute value, and then perform a min-max normalization on the mean to obtain a soft mask).
Because both branches share the same random noise, source-branch prediction errors relative to ground truth can be projected to the target branch, while inter-branch velocity differences also provide latent-level cues of editing-related regions.
The overall process is illustrated in Fig. \ref{fig: framework} (a), where translucent dashed lines denote linear interpolation and bold dashed arrows indicate the target generation.

\subsection{Visual Prompting for Autoregressive Editing}
\label{sec:Visual Prompting}

Text-driven editing typically lacks fine-grained visual control.
The outcome depends on difficult-to-fine-tune natural language prompts and how well the model understands them, yet success is not well ensured.
Controllability and success rate are both critical aspects in video editing \cite{li2025five}.
Since one image is usually easily adjustable by users, we propose regarding the user-provided first frame as a visual prompt by treating it as a previous chunk of the generated video, thereby easily incorporating detailed editing effects users applied to the provided frame into the target generation.
Meanwhile, this approach also enables video editing to leverage existing effective image editing models \cite{labs2025flux1kontextflowmatching, wu2025qwenimagetechnicalreport} by using the model-edited first frame as a visual prompt, thus ensuring easier success.
Such a simple design requires only one additional forward of the video generation model to provide visual prompts, easily making the video editing more desirable.

\section{Experiments}

\subsection{Experimental Setups}

\noindent\textbf{Implementation.}
We build StreamEdit on two pre-trained streaming video generation models: Self Forcing $\text{(SF)}$ \cite{huang2025self} and LongLive $\text{(LL)}$ \cite{yang2025longlive}.
We evaluate both text-driven video editing (Ours $\text{(SF)}$, Ours $\text{(LL)}$) and image-prompted text-driven editing (Ours $\text{(SF)}^\S$, Ours $\text{(LL)}^\S$), where first-frame visual prompts are generated by Qwen-Image-Edit \cite{wu2025qwenimagetechnicalreport}.
For both models, we apply the self-attention bridge and cross-attention boosting to all attention layers, and perform grounding on the first 20 of 30 cross-attention layers.
We fix $\rho=2$ in all evaluations, and set $\omega=4$ for Ours (SF) and $\omega=2$ for the other settings.
We use 15 steps for standard comparisons and 5 steps for long-video streaming editing.
A uniform timestep scheduler is adopted.
Classifier-free guidance is disabled; therefore, the number of function evaluations is 2 $\times$ (steps + 1), consistent with two-sample generation.
All experiments are conducted on a single NVIDIA A100 GPU.

\noindent\textbf{Baselines.}
We focus on text-driven video editing and compare against diverse state-of-the-art methods: image-model-based (TokenFlow \cite{geyer2024tokenflow}, VidToMe \cite{li2024vidtome}, VideoGrain \cite{yang2025videograin}), video-diffusion-based (DMT \cite{yatim2024space}), and flow-matching-based (Pyramid-Edit \cite{li2025five}, Wan-Edit \cite{li2025five}, UniEdit-Flow \cite{jiao2025uniedit}) methods.
For long-video settings, we additionally compare with StreamV2V \cite{liang2024looking} and AdaFlow \cite{zhang2025adaflow}.

\noindent\textbf{Benchmark and Metrics.}
We primarily evaluate on FiVE-Bench \cite{li2025five}, a fine-grained video editing benchmark with 100 collected videos and 420 editing prompt pairs across six edit types: color alteration, material modification, object substitution with/without non-rigid deformation, add, and remove.
FiVE-Bench reports five metric groups: structure/background preservation (structure distance \cite{tumanyan2022splicing}, PSNR, LPIPS \cite{zhang2018unreasonable}, MSE, SSIM \cite{wang2004image}), text alignment (CLIP score \cite{radford2021learning, wu2021godiva}, edit-region CLIP score), image quality (NIQE \cite{saad2012blind}), temporal consistency (motion fidelity \cite{yatim2024space}), and VLM-based editing success (FiVE-Acc).
For baselines already evaluated on FiVE-Bench, we report official numbers obtained on one H100 GPU.
For UniEdit-Flow, we run evaluation using its official implementation.
For long-video evaluation, we collect 30-second videos at 16 FPS from the internet and build 21 editing cases for user study.
The user study evaluates three aspects: overall video quality, background consistency, and editing fidelity.
We report method rankings for each aspect.

\newcommand{\FiveSD}[1]{\colorizeinverse{10.27}{71.36}{#1}}
\newcommand{\FivePSNR}[1]{\colorize{14.71}{28.47}{#1}}
\newcommand{\FiveLPIPS}[1]{\colorizeinverse{49.84}{404.60}{#1}}
\newcommand{\FiveMSE}[1]{\colorizeinverse{21.88}{372.78}{#1}}
\newcommand{\FiveSSIM}[1]{\colorize{51.64}{87.52}{#1}}
\newcommand{\FiveCLIPS}[1]{\colorize{25.69}{27.32}{#1}}
\newcommand{\FiveCLIPSE}[1]{\colorize{20.20}{22.13}{#1}}
\newcommand{\FiveIQA}[1]{\colorizeinverse{4.01}{6.54}{#1}}
\newcommand{\FiveTC}[1]{\colorize{80.59}{90.93}{#1}}
\newcommand{\FiveACC}[1]{\colorize{26.77}{61.19}{#1}}
\newcommand{\FiveTime}[1]{\colorizeinverseLog{0.6}{27.12}{#1}}

\begin{table*}[t]
    \centering
    \caption{\textbf{Comparison of video editing methods} on FiVE-Bench. $^*$ and $\dagger$ denote methods that require optimization and depth/segmentation maps, respectively. $^\S$ indicates methods that using pre-provided edited first frame as additional input (the first frame gathering time is not included in the Time Per Frame). The best and second best results are \textbf{bolded} and \underline{underlined}, with cells highlighted from \colorbox{mylow}{worse} to \colorbox{myhigh}{better}.}
    \resizebox{0.99\textwidth}{!}{
    \begin{tabular}{l|c|cccc|cc|c|c|c|c}
        \toprule
        \multirow{2}{*}{\textbf{Methods}} & \textbf{{Struc.}} &\multicolumn{4}{c|}{\textbf{Background Preservation}} &\multicolumn{2}{c|}{\textbf{Text Alignment}} &\multicolumn{1}{c|}{\textbf{IQA}} &\multicolumn{1}{c|}{\textbf{Temp.}} &\textbf{FiVE} &\textbf{Time (s)} \\
        &Dist.$_{\times 10^3}$$\downarrow$ &PSNR$\uparrow$ &LPIPS$_{\times 10^3}$$\downarrow$ &MSE$_{\times 10^4}$$\downarrow$ &SSIM$_{\times 10^2}$$\uparrow$ &CLIPS.$\uparrow$ &CLIPS.$_{\text{edit}}\uparrow$ &NIQE$\downarrow$ & C.$_{\times 10^2}$$\uparrow$ &Acc$\uparrow$ &Per Frame$\downarrow$ \\
        \midrule
        TokenFlow~\cite{geyer2024tokenflow}  & \FiveSD{35.62} & \FivePSNR{19.06} & \FiveLPIPS{263.61} & \FiveMSE{138.65} & \FiveSSIM{72.51} & \FiveCLIPS{26.46} & \FiveCLIPSE{21.15} &\textbf{\FiveIQA{4.01}} &\FiveTC{89.00} &\FiveACC{27.43} &\FiveTime{8.04} \\
        DMT$^*$~\cite{yatim2024space}  &\FiveSD{85.95} & \FivePSNR{14.71} & \FiveLPIPS{404.60} & \FiveMSE{372.78} & \FiveSSIM{51.64} & \FiveCLIPS{26.66} & \FiveCLIPSE{21.44} &\FiveIQA{5.24} &\FiveTC{82.30} &\FiveACC{48.42} &\FiveTime{25.98} \\
        VidToMe~\cite{li2024vidtome}  &\FiveSD{22.37} & \FivePSNR{21.15} & \FiveLPIPS{263.91} & \FiveMSE{88.75} & \FiveSSIM{70.69} & \FiveCLIPS{26.84} & \FiveCLIPSE{21.05} &\FiveIQA{4.68} &\underline{\FiveTC{90.06}} &\FiveACC{26.77} &\FiveTime{3.25} \\
        VideoGrain$^{\dagger}$~\cite{yang2025videograin} &\underline{\FiveSD{12.40}} &\underline{\FivePSNR{27.05}} & \FiveLPIPS{185.21} & \underline{\FiveMSE{25.10}} & \FiveSSIM{79.13} & \FiveCLIPS{25.69} & \FiveCLIPSE{20.31} &\underline{\FiveIQA{4.08}} &\FiveTC{88.57} &\FiveACC{37.23} &\FiveTime{27.12} \\
        Pyramid-Edit \cite{li2025five}  &\FiveSD{28.65} & \FivePSNR{20.84} & \FiveLPIPS{276.59} & \FiveMSE{95.63} & \FiveSSIM{71.72} & \FiveCLIPS{26.82} & \FiveCLIPSE{20.20} &\FiveIQA{5.48} &\FiveTC{80.59} &\FiveACC{43.84} &\FiveTime{1.44} \\
        Wan-Edit \cite{li2025five}      & \FiveSD{12.53} & \FivePSNR{25.57} & \FiveLPIPS{94.61} & \FiveMSE{41.84} & \FiveSSIM{82.55} & \FiveCLIPS{26.39} & \FiveCLIPSE{21.23} &\FiveIQA{6.54} & \FiveTC{89.43} &\FiveACC{46.97}  &\FiveTime{3.07} \\
        UniEdit-Flow \cite{jiao2025uniedit} & \FiveSD{18.31} & \FivePSNR{24.43} & \FiveLPIPS{223.07} & \FiveMSE{45.23} & \FiveSSIM{78.60} & \FiveCLIPS{26.06} & \FiveCLIPSE{20.92} &\FiveIQA{4.30} & \FiveTC{88.57} &\FiveACC{50.10}  &\FiveTime{1.11} \\ 
        
        \midrule
        \textbf{Ours} (\textbf{SF})  &\textbf{\FiveSD{10.27}} & \textbf{\FivePSNR{28.47}} & \textbf{\FiveLPIPS{49.84}} & \textbf{\FiveMSE{21.88}} & \textbf{\FiveSSIM{87.52}} & \FiveCLIPS{26.95} & \FiveCLIPSE{21.73} &\FiveIQA{4.38} &\textbf{\FiveTC{90.93}} &\FiveACC{51.94} &\textbf{\FiveTime{0.60}} \\
        \textbf{Ours} (\textbf{LL})  &\FiveSD{15.77} & \FivePSNR{25.43} & \FiveLPIPS{65.35} & \FiveMSE{44.36} & \FiveSSIM{84.15} & \textbf{\FiveCLIPS{27.32}} & \textbf{\FiveCLIPSE{22.13}} &\FiveIQA{4.41} &\FiveTC{89.47} &\FiveACC{55.04} &\underline{\FiveTime{0.69}} \\
        \midrule
        \textbf{Ours} (\textbf{SF})$^\S$  &\FiveSD{13.09} & \FivePSNR{26.85} & \underline{\FiveLPIPS{65.32}} & \FiveMSE{37.58} & \underline{\FiveSSIM{85.36}} & \FiveCLIPS{26.99} & \FiveCLIPSE{21.64} &\FiveIQA{4.57} &\FiveTC{89.91} &\underline{\FiveACC{58.64}} &\FiveTime{0.71}$^\S$ \\
        \textbf{Ours} (\textbf{LL})$^\S$  &\FiveSD{15.69} & \FivePSNR{25.74} & \FiveLPIPS{67.04} & \FiveMSE{40.70} & \FiveSSIM{84.29} & \underline{\FiveCLIPS{27.22}} & \underline{\FiveCLIPSE{22.11}} &\FiveIQA{4.61} &\FiveTC{89.12} &\textbf{\FiveACC{61.19}} &\FiveTime{0.76}$^\S$ \\
        
        \bottomrule
    \end{tabular}}
    \label{tab:five}
\end{table*}

\begin{figure*}[t]
    \centering
    \includegraphics[width=0.99\linewidth]{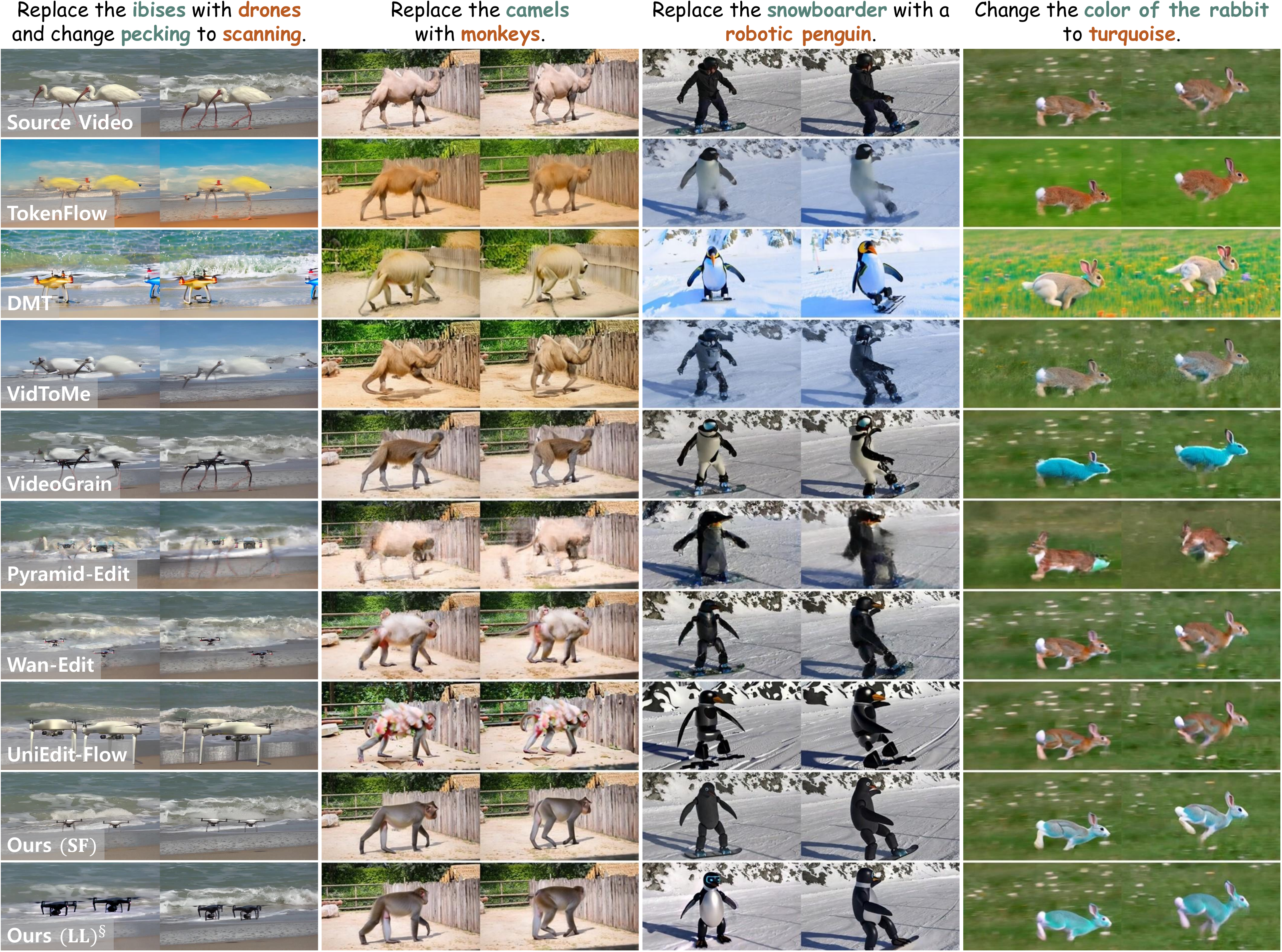}
    \caption{
    \textbf{Qualitative comparisons}.
    We show text-only StreamEdit with Self Forcing ($\text{{Ours} (SF)}$) and image-conditioned StreamEdit with LongLive ($\text{{Ours} (LL)}^{\S}$), demonstrating clear advantages over prior state-of-the-art methods.
    }
    \label{fig: Qualitative comparisons}
\end{figure*}

\subsection{Comparison with State-of-the-Art Methods}

\noindent\textbf{Quantitative Results.}
Tab. \ref{tab:five} reports quantitative comparisons on FiVE-Bench.
Compared with prior methods, StreamEdit---both text-driven and visual-prompted variants---achieves strong gains in text alignment while better preserving structure and context.
Superior LPIPS and SSIM indicate improved photorealism and structural consistency.
The combined results on image quality, motion fidelity, and editing accuracy further show that StreamEdit is both effective and reliable.
Notably, StreamEdit performs strongly on both Self Forcing and LongLive, demonstrating good generalizability.
Self Forcing favors structural/context preservation, while LongLive provides stronger editing effects, offering flexible choices for different user preferences.
Overall, StreamEdit delivers consistently better results with minimal time cost, marking a clear advance over existing methods.

\noindent\textbf{Qualitative Results.}
Fig. \ref{fig: Qualitative comparisons} compares StreamEdit with prior methods.
On challenging cases, advanced flow-based editors (Pyramid-Edit, Wan-Edit, UniEdit-Flow) often fail, especially in color edits (\eg, 4th column), and may produce low-quality, unfinished-looking videos.
In contrast, StreamEdit achieves the intended edits more reliably (\eg, 1st and 2nd columns).
Methods based on video diffusion or image models also struggle to balance editing success with background preservation.
StreamEdit maintains stronger structural/background consistency with the source video while preserving editing fidelity (\eg, 3rd and 4th columns).

\subsection{Ablation Study}

\begin{figure}[t]
\begin{minipage}[h]{0.63\textwidth}
    \centering
    \captionof{table}{\textbf{Quantitative ablation} of the proposed main components.
    S.O.G. indicates the source-oriented guidance and S.A.B. denotes the self-attention bridge.}
    \vspace{1mm}
    \resizebox{1\textwidth}{!}{
    \begin{tabular}{l|ccc|ccc}
        \toprule
        \multirow{2}{*}{\textbf{Methods}} &\multicolumn{3}{c|}{\textbf{Background Preservation}} &\multicolumn{3}{c}{\textbf{Editing Effectiveness}} \\
        &PSNR$\uparrow$ &LPIPS$_{\times 10^3}$$\downarrow$ &SSIM$_{\times 10^2}$$\uparrow$ &CLIPS.$\uparrow$ &CLIPS.$_{\text{edit}}\uparrow$ &FiVE$_\text{Acc}\uparrow$ \\
        \midrule
        w/o S.O.G. (SF) & 21.86 & 122.43 & 72.40 & 27.13 & 21.99 & 53.58 \\
        w/o S.A.B. (SF) & 13.42 & 381.19 & 47.70 & 27.46 & 21.67 & 67.74 \\
        \rowcolor{myhigh!50} \textbf{Ours} (\textbf{SF}) & 28.47 & 49.84 & 87.52 & 26.95 & 21.73 &51.94 \\
        \bottomrule
    \end{tabular}}
    \label{tab:abla_component}
\end{minipage}
\hfill
\begin{minipage}[h]{0.34\textwidth}
    \centering
    \includegraphics[width=\linewidth]{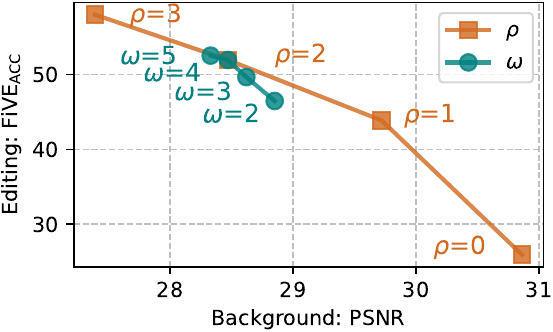}
    \vspace{-0.7cm}
    \captionof{figure}{\textbf{Trade-off} of $\rho$ and $\omega$.}
    \label{fig: curve abla}
    \vspace{-0.6cm}
\end{minipage}
\end{figure}

\begin{figure*}[t]
    \centering
    \includegraphics[width=0.99\linewidth]{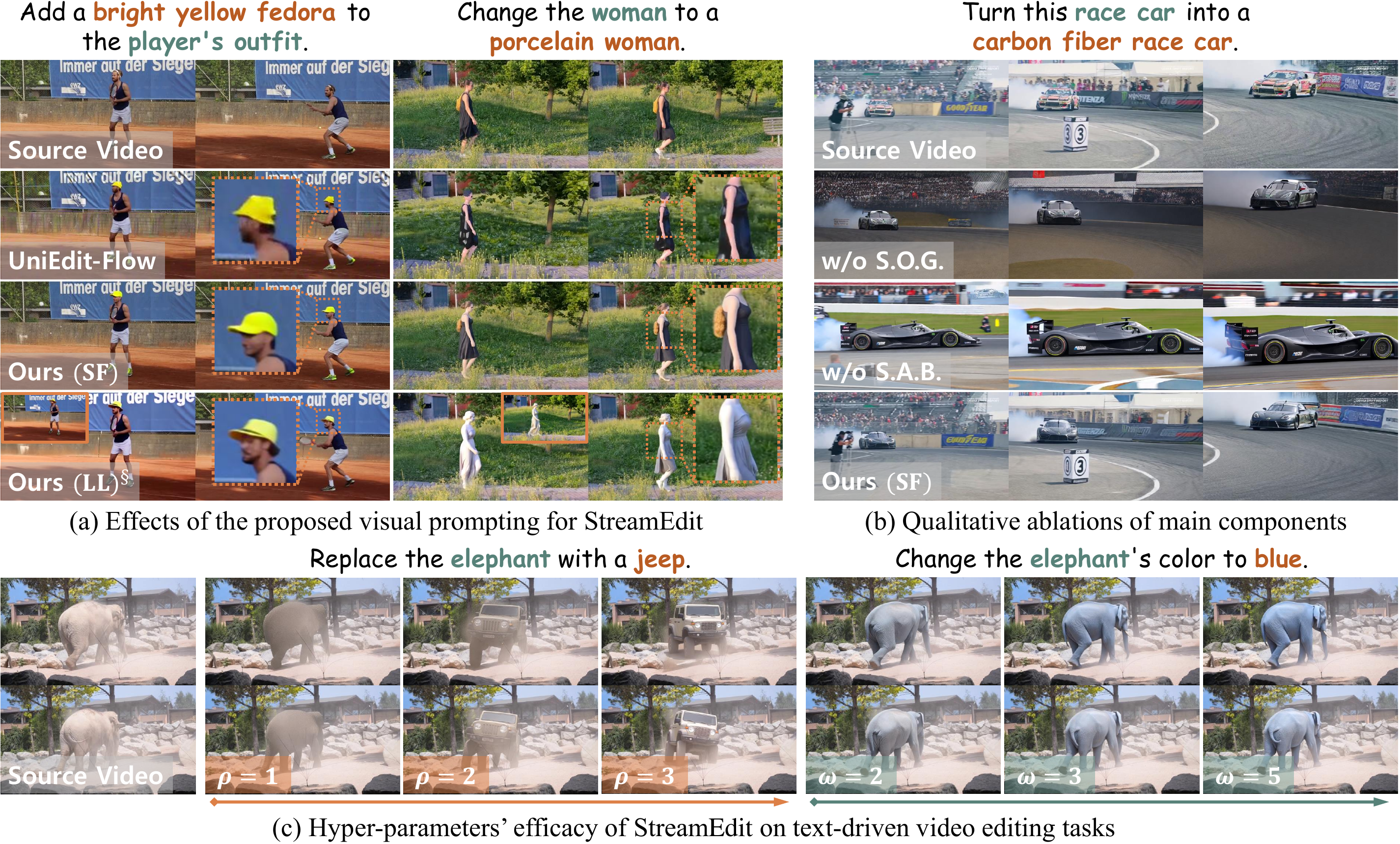}
    \caption{
    \textbf{Qualitative ablation studies} of the proposed StreamEdit.
    }
    \label{fig: ablation studies}
\end{figure*}

\noindent\textbf{Main Components.}
As shown in Tab. \ref{tab:abla_component}, we ablate the two main components of StreamEdit on Self Forcing: source-oriented guidance in stochastic dual-branch sampling (S.O.G.) and the self-attention bridge (S.A.B.).
Source-oriented guidance compensates stochastic errors; removing it may slightly increase edit freedom but notably weakens stable background alignment with the source video (2nd row in Fig. \ref{fig: ablation studies} (b)).
The self-attention bridge is the key mechanism for injecting source conditions into target generation.
Without it, generation becomes weakly constrained by the source and drifts toward the target prompt (2nd row in Tab. \ref{tab:abla_component}), producing outputs largely unrelated to the source video (3rd row in Fig. \ref{fig: ablation studies} (b)).
Together, these two components convert stochastic generation into controllable video-conditioned generation, enabling fast and effective video editing (4th row in Fig. \ref{fig: ablation studies} (b)).

\noindent\textbf{Hyper-parameter Efficacy.}
The two tunable hyper-parameters, the self-attention blending ratio $\rho$ and the cross-attention boosting scale $\omega$, act as control knobs for structural preservation and editing strength, respectively.
The left panel of Fig. \ref{fig: ablation studies} (c) shows the role of $\rho$: smaller $\rho$ improves structural consistency but can weaken edits (no jeep at $\rho=1$), whereas larger $\rho$ allows more flexible edits and better handles complex cases (successful jeep at $\rho=3$).
The right panel of Fig. \ref{fig: ablation studies} (c) shows that larger $\omega$ increases edit intensity (\eg, the elephant becomes bluer as $\omega$ increases).
Fig. \ref{fig: curve abla} further shows that tuning these hyper-parameters yields clear trade-off curves, demonstrating robustness and controllability.

\noindent\textbf{Visual Prompting for video editing.}
To further illustrate visual prompting, we provide representative examples in Fig. \ref{fig: ablation studies} (a).
Its first role is fine-grained visual control over the editing target.
As shown in the left example, text-only editing can be ambiguous across methods/models: for a target fedora, the model may generate a round cap or baseball cap instead.
With an explicit visual prompt (orange solid box), StreamEdit follows the intended target much more reliably.
Its second role is to support difficult edits.
The right example shows a challenging transformation of a woman into a porcelain figure.
Without visual prompting, methods often only smooth the skin.
With a specific visual target, StreamEdit successfully realizes the intended edit.

\subsection{Analyses of the Editing Process}

\begin{figure*}[t]
    \centering
    \includegraphics[width=0.99\linewidth]{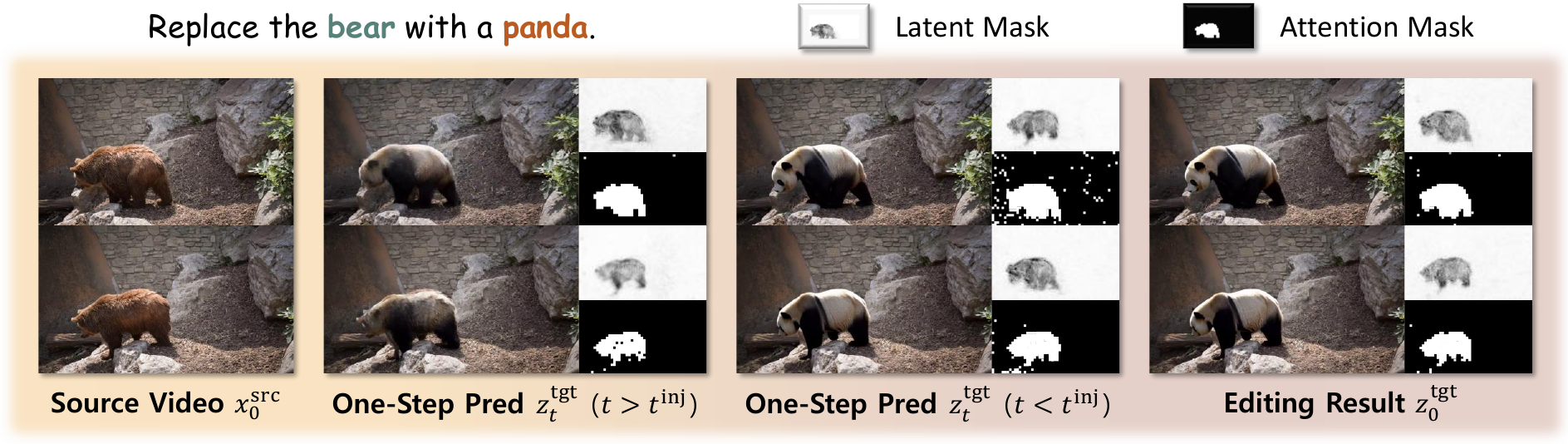}
    \caption{
    \textbf{Visualization of the editing process}.
    We present the editing process from the source video to one-step predictions at $t > t^\text{inj}$ and $t < t^\text{inj}$ and finally to the editing result.
    We visualize the corresponding latent mask and attention mask at the upperright and lower right of each frame.
    Latent masks dynamically update using current timestep's velocity predictions.
    Attention masks change from $M^\text{src}_\text{curr}$ to the union mask $M_\text{curr}$ then to the stored mask $M_\text{prev}$ in the figure.
    }
    \label{fig: exp process}
\end{figure*}

Fig. \ref{fig: exp process} visualizes the editing process of our proposed StreamEdit.
Few-step generation first yields target semantics and then gradually refines them.
Our approach works in a similar way. 
Thus, benefiting from the generation-style paradigm, the target concept rapidly takes shape in the early stages ($z_t^\text{tgt}~(t>t^\text{inj})$).
After injecting source KV, video details are gradually refined towards precise editing ($z_t^\text{tgt}~(t<t^\text{inj})$), and finally yield stable and satisfactory results ($z_0^\text{tgt}$).

In terms of the masks, the upper right of each frame shows the latent masks that are dynamically updated at each timestep to fit the current target concepts. 
Since the denoising process performs differently at different stage \cite{yang2026stable}, this continuous updating ensures more stable and smoother results.
The lower right of each frame indicates its corresponding attention mask.
Attention masks are updated threefold.
Before denoising, we first forward the source clean data to obtain the source mask $M^\text{src}_\text{curr}$ while caching the source KV as the next-chunk previous condition.
At $t = t^\text{inj}$, we extract the target mask $M^\text{tgt}_\text{curr}$ during denoising, and then obtain the union mask $M_\text{curr}$.
Finally after denoising, we forward the target clean data for gaining KV cache while extract a new target mask which will merge with $M^\text{src}_\text{curr}$ to be the previous mask $M_\text{prev}$ for subsquent chunks.
With this design, our masks align accurately with the editing-related regions and require no additional NFEs, thus facilitating precise editing without notable extra cost.

\subsection{Autoregressive Editing for Long Videos}

\begin{figure*}[t]
    \centering
    \includegraphics[width=0.99\linewidth]{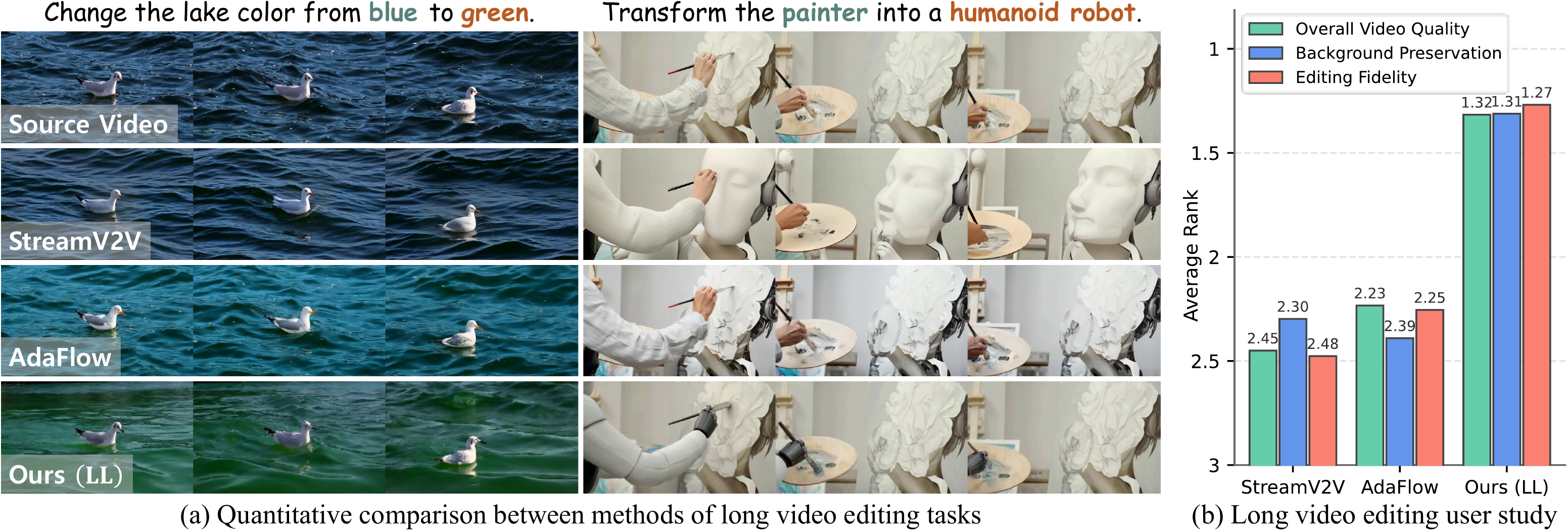}
    \caption{
    \textbf{Comparisons of long video editing} using videos with over 470 frames.
    }
    \label{fig: long result}
\end{figure*}

Fig. \ref{fig: long result} compares StreamEdit with existing long-video editing methods \cite{liang2024looking, zhang2025adaflow}.
We conduct a user study with 20 participants on long-video editing, evaluating overall video quality, background preservation, and editing fidelity.
Our method (Ours (LL), 5 steps) ranks first across all criteria, demonstrating strong effectiveness and reliability for text-driven long-video editing.

\section{Conclusion}

We reconsider video editing as a source-conditioned noise-to-data streaming generation problem, moving beyond the conventional data-to-data paradigm.
Based on this perspective, we propose StreamEdit, which addresses key limitations of existing methods on complex videos, including heavy iterative cost, limited generation quality, and unstable editing performance.
Extensive experiments on text-driven and image-prompted, short- and long-video editing demonstrate the superiority of this paradigm in both effectiveness and efficiency.
In future work, we will further improve scalability and real-time capability, and extend the framework to broader video editing scenarios.

\section*{Acknowledgements}
This work was supported, in part, by the NSERC DG Grant (No. RGPIN-2022-04636), the NSERC Alliance Grant (ALLRP 604621-25), the Vector Institute for AI, and the Canada CIFAR AI Chair program, and a gift fund from Snap Inc.
Resources used in preparing this research were provided, in part, by the Province of Ontario, the Government of Canada through the Digital Research Alliance of Canada \url{alliance.can.ca}, and companies sponsoring the Vector Institute \url{www.vectorinstitute.ai/\#partners}, and Advanced Research Computing at the University of British Columbia. 
Additional resource was provided by the Canada Foundation for Innovation (CFI) via the John R. Evans Leaders Fund (JELF). 
G.J. is supported by a UBC Four Year Doctoral Fellowship.


%
%
\bibliographystyle{splncs04}
\bibliography{main}

\clearpage

\appendix

\setcounter{table}{0}
\setcounter{figure}{0}
\setcounter{equation}{0}
\setcounter{algorithm}{0}
\renewcommand{\thetable}{\Alph{section}.\arabic{table}}
\renewcommand{\thefigure}{\Alph{section}.\arabic{figure}}
\renewcommand{\theequation}{\Alph{section}.\arabic{equation}}
\renewcommand{\thealgorithm}{\Alph{section}.\arabic{algorithm}}

\section*{Appendix}

\begin{itemize}
    \item[$\bullet$] Sec. \ref{sec:appd_implementation} Implementation Details
    \item[$\bullet$] Sec. \ref{sec:appd_add_abla} Additional Experiments
    \item[$\bullet$] Sec. \ref{sec:appd_analyses} Analyses of Existing Training-Free Editing Paradigms
    \item[$\bullet$] Sec. \ref{sec:appd_detail_streaming} Details of Streaming Video Editing Comparison
    \item[$\bullet$] Sec. \ref{sec:appd_add_qualitative} Additional Qualitative Results
    \item[$\bullet$] Sec. \ref{sec:appd_lim} Limitation and Future Work
\end{itemize}

\section{Implementation Details}
\label{sec:appd_implementation}

\subsection{Model Architecture and Corresponding Designs}

The implementation of our proposed StreamEdit is mainly based on the open-source code of Self Forcing \cite{huang2025self} and LongLive \cite{yang2025longlive}.
These models are trained for a visual resolution of 832$\times$480, so we first resize videos to this resolution during editing and then process them back to their original resolution.
The attention mechanism of these models are built on FlashAttention \cite{dao2022flashattention, dao2023flashattention, shah2024flashattention}.

These models are chunk-level autoregressive models. 
The chunk size is 3 latent frames, which is also corresponding to 12 video frames.
Self Forcing employs a fixed-size KV cache of 21 latent frames, without utilizing a sliding window strategy. 
When extending Self Forcing to long video generation or editing, the video clip rolling strategy should be adopted.
LongLive develops a short-window attention with frame-sink tokens, comprising a 9-latent-frame local attention and a 3-latent-frame first-chunk attention sink.
Therefore, LongLive can be directly utilized for long video generation and editing without requiring sampling extension strategies.
Moreover, our proposed visual prompting directly allocates the KV of the condition image to the first position of KV cache on Self Forcing, while three-time-repeatedly filling up the attention sink on LongLive to enhance visual conditioning.

\subsection{Implementation of Cross-Attention Boosting}

Since the FlashAttention does not provide explicit score maps, and replacing FlashAttention with general matrix multiplication would significantly reduce computational efficiency, the proposed cross-attention boosting cannot be directly achieved through simple addition.
Therefore, we design an indirect method for cross-attention boosting without altering computational complexity by incorporating additional FlashAttention operations.
Specifically, FlashAttention directly returns:
\begin{equation}
    \mathtt{FlashAttention}(Q,K,V) = \frac{\sum_p \exp(S_{p,q}) V_p}{\sum_p \exp(S_{p,q})},
\end{equation}
where $S_{p,q}$ is the dot product score of text token $p$ and visual query $q$.
We decompose the unmasked cross-attention boosting into a numerator-denominator pair, computing each separately via FlashAttention:
\begin{equation}
    \begin{split}
        &a_{\text{num}} = \mathtt{FlashAttention}(Q,K,wV) = \frac{\sum_p \exp(S_{p,q}) (w_p V_p)}{\sum_p \exp(S_{p,q})}, \\
        &a_{\text{denom}} = \mathtt{FlashAttention}(Q,K,w) = \frac{\sum_p \exp(S_{p,q}) w_p}{\sum_p \exp(S_{p,q})}, 
    \end{split}
\end{equation}
where $w_p = \omega$ if $p \in \mathcal{T}$ else 1.
Subsequently, by performing the fraction calculation, we can obtain the desired output after unmasked cross-attention boosting:
\begin{equation}
    \frac{a_{\text{num}}}{a_{\text{denom}}} = \frac{\frac{\sum_p \exp(S_{p,q}) w_p V_p}{\sum_p \exp(S_{p,q})}}{\frac{\sum_p \exp(S_{p,q}) w_p}{\sum_p \exp(S_{p,q})}} = 
    \frac{\sum_p w_p \exp(S_{p,q}) V_p}{\sum_p w_p \exp(S_{p,q})} = 
    \frac{\sum_p \exp(S_{p,q} + \ln w_p ) V_p}{\sum_p \exp(S_{p,q}+ \ln w_p )} .
\end{equation}
Subsequently, we blend it with the original cross-attention via the grounding mask, applying the boosted result to editing-related regions while preserving the original values in editing-irrelevant regions.
Finally, we achieve trigger- and region-aware editing enhancement while retaining FlashAttention operations to avoid significantly increasing computational complexity.

\subsection{Overall Algorithm}

\begin{algorithm}[!ht]
\caption{\textbf{StreamEdit Inference}}
\label{alg:StreamEdit}
\begin{algorithmic}[1]
\Require $C$-chunked source video $\{x_{0,c}^{\mathrm{src}}\}_{c=1}^{C}$
\Require Prompts $\mathcal{P}^{\mathrm{src}}$, $\mathcal{P}^{\mathrm{tgt}}$, trigger sets $\mathcal{T}^\mathrm{src}, \mathcal{T}^\mathrm{tgt}$, (optional) edited first frame $f^\text{tgt}$
\Require Model, setting, and tunable hyper-parameters $v_\theta$, $t^{\mathrm{inj}}$, $\rho,\omega$
\Require Timesteps $\{t_i\}_{i=0}^{N}$
\Ensure Target edited video $\{z_{0,c}^{\mathrm{tgt}}\}_{c=1}^{C}$

\State $(K^{\mathrm{src}}_{\mathrm{prev}},V^{\mathrm{src}}_{\mathrm{prev}})\gets \varnothing$, $(K^{\mathrm{tgt}}_{\mathrm{prev}},V^{\mathrm{tgt}}_{\mathrm{prev}})\gets \varnothing$, $M_{\mathrm{prev}}\gets \varnothing$
\If{$f^\text{tgt}$ is provided}
    \State $f^\text{src} \gets $ the first frame of source video
    \State $M^\mathrm{src}, (K^\mathrm{src}_{0}, V^\mathrm{src}_{0}) \gets \Call{Ground}{v_\theta, f^\text{src}, 0, \mathcal{P}^\mathrm{src}, \mathcal{T}^\mathrm{src}}$
    \State $M^\mathrm{tgt}, (K^\mathrm{tgt}_{0}, V^\mathrm{tgt}_{0}) \gets \Call{Ground}{v_\theta, f^\text{tgt}, 0, \mathcal{P}^\mathrm{tgt}, \mathcal{T}^\mathrm{tgt}}$
    \State $(K^{\mathrm{src}}_{\mathrm{prev}},V^{\mathrm{src}}_{\mathrm{prev}}) \mathtt{.append}(K^\mathrm{src}_{0}, V^\mathrm{src}_{0})$, $(K^{\mathrm{tgt}}_{\mathrm{prev}},V^{\mathrm{tgt}}_{\mathrm{prev}})\mathtt{.append}(K^\mathrm{tgt}_{0}, V^\mathrm{tgt}_{0})$
    \State $M_{\mathrm{prev}}\mathtt{.append}(M^\mathrm{src} \cup M^\mathrm{tgt})$
\EndIf
\For{$c=1$ to $C$}
    \State $x_{0}^{\mathrm{src}} \gets x_{0,c}^{\mathrm{src}}$, $z_{t_N}^{\mathrm{tgt}}\gets x_{0,c}^{\mathrm{src}}$
    \State $M^{\mathrm{src}}, (K^{\mathrm{src}}_{\mathrm{c}},V^{\mathrm{src}}_{\mathrm{c}}) \gets \Call{Ground}{v_\theta, x_{0}^{\mathrm{src}},0,\mathcal{P}^{\mathrm{src}},\mathcal{T}^\mathrm{src}}$
    \State $M_{\mathrm{curr}} \gets M^\mathrm{src}$
    \For{$i=N$ downto $1$}
        \State Sample $\epsilon_{t_i}\sim\mathcal{N}(0,I)$
        \State $x_{t_i}^{\mathrm{src}}\gets (1-t_i)x_{0}^{\mathrm{src}}+t_i\epsilon_{t_i}$,
        $x_{t_i}^{\mathrm{tgt}}\gets (1-t_i)z_{t_i}^{\mathrm{tgt}}+t_i\epsilon_{t_i}$
        \State $v_{t_i}^{\mathrm{src}}, (Q,K,V)
            \gets \textproc{BridgeSA}\!\left(
            \begin{aligned}
            &v_\theta, x_{t_i}^{\mathrm{src}}, x_{t_i}^{\mathrm{tgt}},
            K^{\mathrm{src}}_{\mathrm{prev}}, V^{\mathrm{src}}_{\mathrm{prev}},\\
            &K^{\mathrm{tgt}}_{\mathrm{prev}}, V^{\mathrm{tgt}}_{\mathrm{prev}},
            M_{\mathrm{prev}}, M_{\mathrm{curr}}, \rho, t_i
            \end{aligned}
            \right)$
        \If{$t_i=t^{\mathrm{inj}}$}
            \State $M^{\mathrm{tgt}}, \_ \gets \Call{Ground}{v_\theta, x_{t_i}^{\mathrm{tgt}}, t_i,\mathcal{P}^{\mathrm{tgt}},\mathcal{T}^\mathrm{tgt}}$ ~~~ {\color{gray!60} \% \textit{Sync with} $\textproc{BridgeSA}$}
            \State $M_\mathrm{curr}\gets M^{\mathrm{src}}\cup M^{\mathrm{tgt}}$
        \EndIf
        \State $A_{\mathrm{CA}}^{\mathrm{tgt}}\gets \Call{BoostCA}{A_{\mathrm{CA}}^{\mathrm{tgt}},M_\mathrm{curr},\mathcal{P}^\mathrm{tgt},\mathcal{T}^\mathrm{tgt},\omega}$ ~~~~ {\color{gray!60} \% \textit{Sync with} $\textproc{BridgeSA}$}
        \State $v_{t_i}^{\mathrm{tgt}}\gets v_\theta^{*}(x_{t_i}^{\mathrm{tgt}},t_i\mid x_{t_i}^{\mathrm{src}};Q,K,V,A_{\mathrm{CA}}^{\mathrm{tgt}})$ ~~~~~~~~~~~~~~ {\color{gray!60} \% \textit{Sync with} $\textproc{BridgeSA}$}
        \State $\tilde{v}_{t_i}^{\mathrm{tgt}}\gets \Call{SOG}{v_{t_i}^{\mathrm{tgt}},v_{t_i}^{\mathrm{src}},x_{0}^{\mathrm{src}},\epsilon_{t_i}}$
        \State $z_{t_{i-1}}^{\mathrm{tgt}}\gets x_{t_i}^{\mathrm{tgt}}-t_i\tilde{v}_{t_i}^{\mathrm{tgt}}$
    \EndFor
    \State $z_{0,c}^{\mathrm{tgt}}\gets z_{t_0}^{\mathrm{tgt}}$
    \State $M^\mathrm{tgt}, (K^{\mathrm{tgt}}_{\mathrm{c}},V^{\mathrm{tgt}}_{\mathrm{c}}) \gets \Call{Ground}{v_\theta, z^\mathrm{tgt}_{0,c}, 0, \mathcal{P}^\mathrm{tgt}, \mathcal{T}^\mathrm{tgt}}$
    \State $(K^{\mathrm{src}}_{\mathrm{prev}},V^{\mathrm{src}}_{\mathrm{prev}}) \mathtt{.append}(K^{\mathrm{src}}_{\mathrm{c}},V^{\mathrm{src}}_{\mathrm{c}}) $, $(K^{\mathrm{tgt}}_{\mathrm{prev}},V^{\mathrm{tgt}}_{\mathrm{prev}})\mathtt{.append}(K^{\mathrm{tgt}}_{\mathrm{c}},V^{\mathrm{tgt}}_{\mathrm{c}})$
    \State $M_{\mathrm{prev}}\mathtt{.append}(M^\mathrm{src} \cup M^\mathrm{tgt})$
\EndFor
\State \Return $\{z_{0,c}^{\mathrm{tgt}}\}_{c=1}^{C}$
\end{algorithmic}
\end{algorithm}

\begin{algorithm}[!ht]
\caption{\textbf{Subroutines of StreamEdit}}
\label{alg:StreamEdit_sub}
\begin{algorithmic}[1]

\Function{Ground}{$v_\theta, x,t,P,\mathcal{T}$}
    \State Get self-attn KV and $l \in \{1, ..., 20\}$ layers' cross-attn $A^l_{\mathrm{CA}}(p,Q_{\mathrm{CA}})$ of $v_\theta(x,t)$
    \State $M_{l} \gets 
    \frac{1}{|\mathcal{T}|}\sum_{p\in\mathcal{T}}A^l_{\mathrm{CA}}(p,Q_{\mathrm{CA}}) 
    -
    \frac{1}{|\mathcal{P}\setminus\mathcal{T}|}\sum_{p\in\mathcal{P}\setminus\mathcal{T}}A^l_{\mathrm{CA}}(p,Q_{\mathrm{CA}})    $
    \State $M \gets H\!\left(\mathtt{Average}(M_{l})\right)$
    \If{model is LongLive and $x$ is the first frame}
        \State $M \gets \mathtt{Triple}(M)$, $(K, V) \gets \mathtt{Triple}(K, V)$  \quad \quad \quad \quad {\color{gray!60} \% \textit{Fill up sink tokens} }
    \EndIf
    \State \Return $M, (K,V)$
\EndFunction

\Function{BridgeSA}{$v_\theta, x^{\mathrm{src}},x^{\mathrm{tgt}},K^{\mathrm{src}}_{\mathrm{prev}},V^{\mathrm{src}}_{\mathrm{prev}},K^{\mathrm{tgt}}_{\mathrm{prev}},V^{\mathrm{tgt}}_{\mathrm{prev}},M_{\mathrm{prev}},M_{\mathrm{curr}},\rho,t_i$}
    \State $r_{t_i}\gets 1-t_{i-1}^{\rho}$
    \For{self-attns in $v_\theta(x^{\mathrm{src}}, t_i)$ and $v_\theta(x^{\mathrm{tgt}}, t_i)$}
        \State Extract $(Q_{\mathrm{SA}}^{\mathrm{src}},K_{\mathrm{SA}}^{\mathrm{src}},V_{\mathrm{SA}}^{\mathrm{src}})$ and $(Q_{\mathrm{SA}}^{\mathrm{tgt}},K_{\mathrm{SA}}^{\mathrm{tgt}},V_{\mathrm{SA}}^{\mathrm{tgt}})$
        \State $\tilde{Q}_{\mathrm{SA}}^{\mathrm{tgt}}\gets r_{t_i}Q_{\mathrm{SA}}^{\mathrm{tgt}}+(1-r_{t_i})Q_{\mathrm{SA}}^{\mathrm{src}}$
        \State $\tilde{K}_{\mathrm{SA}}^{\mathrm{tgt}}\gets r_{t_i}K_{\mathrm{SA}}^{\mathrm{tgt}}+(1-r_{t_i})K_{\mathrm{SA}}^{\mathrm{src}}$
        \State $\tilde{K}_{\mathrm{prev}}^{\mathrm{tgt}}\gets M_{\mathrm{prev}}\odot\!\left(r_{t_i}K_{\mathrm{prev}}^{\mathrm{tgt}}+(1-r_{t_i})K_{\mathrm{prev}}^{\mathrm{src}}\right)+(1-M_{\mathrm{prev}})\odot K_{\mathrm{prev}}^{\mathrm{tgt}}$
        \State self-attn$^\mathrm{src} \gets \mathtt{Attn}(Q_{\mathrm{SA}}^{\mathrm{src}},K_{\mathrm{SA}}^{\mathrm{src}},V_{\mathrm{SA}}^{\mathrm{src}})$
        \State $(Q,K,V)^\mathrm{tgt} \gets 
            \begin{aligned}
            \tilde{Q}_{\mathrm{SA}}^{\mathrm{tgt}}, 
            &K=[\tilde{K}^\text{tgt}_\text{SA}, \tilde{K}^\text{tgt}_\text{prev}, [t_i<t^\text{inj}]\cdot(M_\text{curr} \odot K^\text{src}_\text{SA})], \\
            &V=[V^\text{tgt}_\text{SA}, ~V^\text{tgt}_\text{prev}, ~[t_i<t^\text{inj}]\cdot(M_\text{curr} \odot V^\text{src}_\text{SA})]
            \end{aligned}$
    \EndFor
    \State $v_{t_i}^{\mathrm{src}} \gets v_\theta(x^{\mathrm{src}}, t_i)$
    \State \Return $v_{t_i}^{\mathrm{src}}, (Q,K,V)^\mathrm{tgt}$
\EndFunction

\Function{BoostCA}{$A_{\mathrm{CA}},M,\mathcal{P},\mathcal{T},\omega$}
    \State Obtain $S(p,q)$ from $A_{\mathrm{CA}}$, $p \in \mathcal{P}$, $q$ is visual query
    \State $w_{p,q}\gets \begin{cases}
    \omega,& p\in\mathcal{T}\ \wedge\ M(q)=1\\
    1,& \text{otherwise}
    \end{cases}$
    \State \Return $\displaystyle
    \frac{\exp(S(p,q)+\ln w_{p,q})}
    {\sum_{j}\exp(S(j,q)+\ln w_{j,q})}$
\EndFunction

\Function{SOG}{$v^{\mathrm{tgt}},v^{\mathrm{src}},x_0^{\mathrm{src}},\epsilon_t$}
    \State $g_t\gets (\epsilon_t-x_0^{\mathrm{src}})-v^{\mathrm{src}}$
    \State \Return $v^{\mathrm{tgt}}+\mathrm{AMN}(v^{\mathrm{tgt}}-v^{\mathrm{src}})\odot g_t$
\EndFunction

\end{algorithmic}
\end{algorithm}

To present the proposed pipeline more clearly, we provide the detailed overall algorithm of StreamEdit in Alg. \ref{alg:StreamEdit} with its subroutines in Alg. \ref{alg:StreamEdit_sub}.
The required user provided inputs including the source video, source prompt (\eg, ``\textit{\underline{A golden retriever} wearing a red harness is walking slowly with its nose close to the dry, leaf-covered ground in a fenced yard next to a bush and a road in the background.}''), target prompt (\eg, ``\textit{\underline{A black and white Border Collie} wearing a red harness is walking slowly with its nose close to the dry, leaf-covered ground in a fenced yard next to a bush and a road in the background.}''), source trigger words (\eg, ``\textit{A golden retriever}''), and target trigger words (\eg, ``\textit{A black and white Border Collie}'').
The tunable hyper-parameters consists of the blending ratio $\rho$ and the boosting scale $\omega$.

\subsection{Evaluation Metrics}

In the manuscript, we report the comparison results averaged across six editing types on FiVE-Bench \cite{li2025five}.
Most baseline results reported on FiVE-Bench are directly derived from its official evaluation results.
Besides UniEdit-Flow \cite{jiao2025uniedit}, we follow its official implementation of video editing ($\alpha=0.8$, $\omega=5$, $N$=25) to validate and report results on FiVE-Bench.
Here we explain the specific meaning of each metric.

\noindent \textbf{Structure Distance.} 
The structure distance \cite{tumanyan2022splicing} is the self-similarity of deep spatial features extracted from DINO-ViT and uses cosine similarity between frame features.

\noindent \textbf{Background Preservation.} 
FiVE-Bench porvides a series of precise mask of editing-irrelevant regions.
PSNR, \cite{zhang2018unreasonable}, MSE, and SSIM \cite{wang2004image} are calculated under their standard formulas in editing-irrelevant regions.

\noindent \textbf{Edit Prompt-Image Consistency.} 
The CLIP similarity \cite{radford2021learning, wu2021godiva} is extended to the video data by per-frame validating.
CLIPS. denotes the whole-video CLIP similarity.
CLIPS.$_{\text{edit}}$ indicates the CLIP similarity between target prompt and masked edited video, which demonstrate the regional editing accuracy.

\noindent \textbf{Image Quality Assessment.}
NIQE \cite{saad2012blind} is calculated for image quality assessment evaluation.
It represent the distortion of the input video using its statistical information at frequency domain, thus efficiently providing video quality assessment without reference.

\noindent \textbf{Temporal Consistency.}
A motion fidelity score \cite{yatim2024space} is utilized for temporal consistency evaluation.
It extracts tracklets from both the original and generated videos using a pre-trained tracker model. 
Then, drawing inspiration from Chamfer distance, it measures whether the two videos retain the same overall motion patterns by employing bidirectional nearest-neighbor trajectory displacement correlation.

\noindent \textbf{Editing Accuracy.}
The reported FiVE-Acc is the average of FiVE-YN, FiVE-MC, FiVE-$\cup$, and FiVE-$\cap$.
FiVE-YN determines whether editing is successful based on yes/no questions, requiring the model to simultaneously identify that the original object has disappeared and the target object has appeared.
FiVE-MC uses a multiple-choice question format to determine whether the edited video resembles the target object or the original object more closely, thereby measuring the accuracy of target object recognition.
FiVE-$\cup$ indicates that success is achieved if either FiVE-YN or FiVE-MC is successful, reflecting a more relaxed overall editing success rate.
FiVE-$\cap$ indicates that both FiVE-YN and FiVE-MC must be satisfied simultaneously for success, reflecting a stricter measure of high-quality editing success rate.

\section{Additional Experiments}
\label{sec:appd_add_abla}

\subsection{Cross-Attention Boosting}

\begin{table*}[t]
    \centering
    \caption{\textbf{Ablations of cross-attention boosting}. 
    C.A.B. indicates the proposed cross-attention boosting.
    $^\S$ indicates methods that using pre-provided edited first frame as additional input.
    \ding{55} denotes using the original cross-attention without boosting.}
    \resizebox{0.99\textwidth}{!}{
    \begin{tabular}{l|c|c|cccc|cc|c|c|c}
        \toprule
        \multirow{2}{*}{\textbf{Methods}} & \multirow{2}{*}{\textbf{C.A.B.}} & \textbf{{Struc.}} &\multicolumn{4}{c|}{\textbf{Background Preservation}} &\multicolumn{2}{c|}{\textbf{Text Alignment}} &\multicolumn{1}{c|}{\textbf{IQA}} &\multicolumn{1}{c|}{\textbf{Temp.}} &\textbf{FiVE} \\
        & &Dist.$_{\times 10^3}$$\downarrow$ &PSNR$\uparrow$ &LPIPS$_{\times 10^3}$$\downarrow$ &MSE$_{\times 10^4}$$\downarrow$ &SSIM$_{\times 10^2}$$\uparrow$ &CLIPS.$\uparrow$ &CLIPS.$_{\text{edit}}\uparrow$ &NIQE$\downarrow$ & C.$_{\times 10^2}$$\uparrow$ &Acc$\uparrow$ \\
        \midrule
        \textbf{Ours} (\textbf{SF}) & \ding{55} & 8.57 & 28.99 & 47.08 & 18.91 & 88.01 & 26.36 & 21.29 & 4.39 & 91.20 & 39.05 \\
        \textbf{Ours} (\textbf{LL}) & \ding{55} & 14.26 & 25.74 & 63.75 & 41.92 & 84.43 & 26.93 & 21.87 & 4.44 & 89.75 & 46.46 \\
        \textbf{Ours} (\textbf{SF})$^\S$ & \ding{55} & 12.97 & 26.91 & 65.02 & 37.29 & 85.41 & 26.88 & 21.56 & 4.57 & 89.93 & 56.29 \\
        \textbf{Ours} (\textbf{LL})$^\S$ & \ding{55} & 15.45 & 25.80 & 66.72 & 40.13 & 84.39 & 27.12 & 22.04 & 4.62 & 89.35 & 54.78 \\
        \midrule
        \rowcolor{myhigh!50}
        \textbf{Ours} (\textbf{SF}) & 4 &{{10.27}} & {{28.47}} & {{49.84}} & {{21.88}} & {{87.52}} & {26.95} & {21.73} &{4.38} &{{90.93}} &{51.94} \\
        \rowcolor{myhigh!50}
        \textbf{Ours} (\textbf{LL}) & 2 &{15.77} & {25.43} & {65.35} & {44.36} & {84.15} & {{27.32}} & {{22.13}} &{4.41} &{89.47} &{55.04} \\
        \rowcolor{myhigh!50}
        \textbf{Ours} (\textbf{SF})$^\S$ & 2 &{13.09} & {26.85} & {{65.32}} & {{37.58}} & {{85.36}} & {26.99} & {21.64} &{4.57} &{89.91} &{{58.64}} \\
        \rowcolor{myhigh!50}
        \textbf{Ours} (\textbf{LL})$^\S$ & 2 &{15.69} & {25.74} & {67.04} & {40.70} & {84.29} & {{27.22}} & {{22.11}} &{4.61} &{89.12} &{{61.19}} \\
        
        \bottomrule
    \end{tabular}}
    \label{tab:cab}
\end{table*}

Tab. \ref{tab:cab} presents the results after dropping the proposed cross-attention boosting.
Without employing cross-attention boosting, our proposed method still achieves successful editing while maintaining high background fidelity and preserving a favorable level of CLIP similarity with the target prompt.
It demonstrates that the method already exhibits outstanding performance after incorporating sampling design and self-attention bridges to meet fundamental paradigm requirements, further highlighting the superiority of the generative paradigm for streaming video editing.
Meanwhile, the reduction in background preservation after introducing cross-attention boosting is negligible, while the improvement in editing success rate (FiVE-Acc) is remarkably significant.
This further demonstrates that cross-attention boosting serves as an excellent valve, enabling flexible control over editing strength while causing virtually no visible impact on editing-irrelevant regions.

Therefore, while cross-attention boosting is not an essential component of the generative paradigm, it provides a stable and reliable control valve for the specific approach we have designed, offering users greater flexibility and controllability.

\begin{table}[t]
\centering
    \caption{\textbf{Ablations} of $t^\text{inj}$, comparison with training-based methods, and comparison with image-conditioned training-free and -based methods. 
    $^\S$ indicates methods that using pre-provided edited first frame as additional input.
    The image-conditioned methods are re-implemented using their official codes, while the first-frame conditions are all provided by the same Qwen-Image-Edit \cite{wu2025qwenimagetechnicalreport} model across methods.}
\resizebox{0.99\linewidth}{!}{
\begin{tabular}{l|cccc|cc|c|c|c|c}
\toprule
    \multirow{2}{*}{\textbf{Methods}} &\multicolumn{4}{c|}{\textbf{Background Preservation}} &\multicolumn{2}{c|}{\textbf{Text Alignment}} &\multicolumn{1}{c|}{\textbf{IQA}} &\multicolumn{1}{c|}{\textbf{Temp.}} &\textbf{FiVE} &\textbf{Time (s)} \\
    & PSNR$\uparrow$ & LPIPS$_{\times 10^3}$$\downarrow$ &MSE$_{\times 10^4}$$\downarrow$ &SSIM$_{\times 10^2}$$\uparrow$ &CLIPS.$\uparrow$ &CLIPS.$_{\text{edit}}\uparrow$ &NIQE$\downarrow$ & C.$_{\times 10^2}$$\uparrow$ &Acc$\uparrow$ &Per Frame$\downarrow$ \\
\midrule
    Ours (SF), $\omega=2$ & 28.85 & 46.58 & 19.70 & 87.71 & 26.62 & 21.66 & 4.38 & 91.22 & 46.46 & 0.60 \\
    Ours (SF), $\omega=3$ & 28.62 & 48.59 & 20.75 & 87.73 & 26.86 & 21.70 & 4.38 & 91.02 & 49.65 & 0.60 \\
    Ours (SF), $\omega=5$ & 28.33 & 50.56 & 22.75 & 87.39 & 26.97 & 21.81 & 4.38 & 90.89 & 52.52 & 0.60 \\
\midrule
    Ours (SF), $\rho=0$ & 30.86 & 38.56 & 13.09 & 89.21 & 25.74 & 20.55 & 4.39 & 92.18 & 25.92 & 0.60 \\
    Ours (SF), $\rho=1$ & 29.72 & 44.13 & 16.29 & 88.46 & 26.55 & 21.34 & 4.39 & 91.50 & 43.85 & 0.60 \\
    Ours (SF), $\rho=3$ & 27.39 & 54.81 & 27.97 & 86.69 & 27.15 & 21.99 & 4.37 & 90.75 & 57.99 & 0.60 \\
\midrule
    Ours (SF), $t^{\mathrm{inj}}=0.3$ & 28.49 & 49.32 & 21.60 & 87.24 & 26.95 & 21.76 & 4.41 & 90.99 & 51.17 & 0.60 \\
    Ours (SF), $t^{\mathrm{inj}}=0.7$ & 27.43 & 52.60 & 27.89 & 86.57 & 26.96 & 21.87 & 4.33 & 90.51 & 52.10 & 0.60 \\
\midrule
    InsViE \cite{wu2025insvie} & 15.60 & 273.51 & 349.12 & 46.90 & 25.01 & 19.74 & 4.61 & 72.94 & 16.82 & 2.42 \\
    LucyEdit \cite{decart2025lucyedit} & \underline{26.76} & \underline{81.92} & \underline{40.07} & \underline{82.82} & \underline{25.29} & \underline{20.44} & \textbf{4.33} & \underline{89.54} & \underline{28.61} & \underline{1.12} \\
    \rowcolor{myhigh!50}
    \textbf{Ours (SF)} & \textbf{28.47} & \textbf{49.84} & \textbf{21.88} & \textbf{87.52} & \textbf{26.95} & \textbf{21.73} & \underline{4.38} & \textbf{90.93} & \textbf{51.94} & \textbf{0.60} \\
\midrule
    AnyV2V$^{\S}$ \cite{ku2024anyv2v} & 20.35 & 167.36 & 121.71 & 66.04 & 27.15 & \underline{21.63} & 4.68 & 79.54 & 66.27 & 6.65$^{\S}$ \\
    Senorita$^{\S}$ \cite{zi2025senorita} & \underline{23.53} & \underline{109.82} & \underline{70.40} & \underline{79.76} & \underline{27.28} & 21.43 & \textbf{4.59} & \textbf{88.89} & \underline{76.17} & \underline{3.86}$^{\S}$ \\
    \rowcolor{myhigh!50}
    \textbf{Ours (LL)}$^{\S}$, ${\omega=4, \rho=4}$ & \textbf{24.01} & \textbf{79.84} & \textbf{59.13} & \textbf{82.40} & \textbf{27.44} & \textbf{22.13} & \textbf{4.59} & \underline{87.92} & \textbf{78.76} & \textbf{0.76}$^{\S}$ \\
\bottomrule
\end{tabular}%
}
\label{tab:tinj_training}
\end{table}

\subsection{Detailed Hyper-Parameter Ablation Studies}

In addition to the trade-off curve of hyper-parameters $\omega$ and $\rho$ in the main text, we further demonstrate their detailed benchmark evaluation results in Tab. \ref{tab:tinj_training}.
Meanwhile, we show $t^\text{inj}=0.3/0.7$ results to indicate its effect.
Within a reasonable range, this hyper-parameter will not excessively affect overall performance, while $t^\text{inj}=0.5$ in the manuscript is selected for easier understanding.

\subsection{Comparison with Training-based Methods}

Training-based methods have achieved remarkable results in short video editing.
Thus, as shown in Tab. \ref{tab:tinj_training}, we further compare StreamEdit (SF and LL$^{\S}_{\omega=4,\rho=4}$) with existing open-source training-based methods, \ie, InsViE \cite{wu2025insvie}, LucyEdit \cite{decart2025lucyedit}, and Senorita$^{\S}$ \cite{zi2025senorita}.
We re-implement these methods using their publicly released code and models and keep these methods' officially provided settings: step=50 and CFG=7.5 for InsViE, step=50 and CFG=5.0 for LucyEdit, step=30 and CFG=4.0 for Senorita$^{\S}$.
For Ours (LL)$^{\S}$ and Senorita$^{\S}$, we use the same first frame conditions generated by Qwen-Image-Edit for fairness.
The results indicate the superior preservation-editing trade-off of StreamEdit with the lowest time cost.
Our training-free design also shows better generalizability than training-based methods with more generalizable framework, longer edit length, and better trade-off, while it has potentials for training methods in long-video data preparation, distillation supervision, and streaming paradigm reference.

Additionally, the trigger words of the removal editing type in FiVE-Bench \cite{li2025five} are not always consistent and correct, sometimes the source trigger corresponds to the object that need to be removed (\eg, for removing the two dogs, source trigger: ``\textit{two dogs}'', target trigger: ``\textit{without two dogs}''), and sometimes the target corresponds to the removed object (\eg, for removing the red cap, source trigger: ``\textit{a young boy}'', target trigger: ``\textit{a red cap}''). 
Thus, we correct such trigger words of the removal editing type, so that the source triggers correspond correctly to the object that needs to be removed and the target triggers point to the desired result (\eg, for removing the two dogs, source trigger: ``\textit{two dogs}'', target trigger: ``\textit{an open field}'').
These corrected trigger words and prompts will be released.
This modification is only applied in the comparisons of training-based and our methods.
Across all evaluations of our proposed method, only Ours (LL)$^{\S}_{\omega=4,\rho=4}$ used the modified removal editing prompts.
Except for its result in the last row of Tab. \ref{tab:tinj_training}, all validations of StreamEdit in this paper retain the original trigger words and prompts of FiVE-Bench to ensure fairness.

\subsection{Comparison with Image-Conditioned Methods}

To evaluate the effectiveness of visual-prompted StreamEdit, we compare image-based methods with Ours (LL)$^\S$ under the same first frame conditions, including training-free AnyV2V$^\S$ and training-based Senorita$^\S$.
We re-implement AnyV2V$^\S$ using their official code but remove the initial instructPix2Pix \cite{brooks2023instructpix2pix} first frame generation, and directly use the Qwen-Image-Edit-generated first frames as conditions for fairness. 
The results further show StreamEdit's superiority for image-conditioned video editing with the help of the proposed visual prompting.

\subsection{Few-Step Quantitative Results}

\begin{table*}[t]
    \centering
    \caption{\textbf{Ablations of sampling steps}. 
    $^\S$ indicates methods that using pre-provided edited first frame as additional input.
    We validate different variants of our proposed method on FiVE-Bench with few-step settings.}
    \resizebox{0.99\textwidth}{!}{
    \begin{tabular}{l|c|c|cccc|cc|c|c|c}
        \toprule
        \multirow{2}{*}{\textbf{Methods}} & \multirow{2}{*}{\textbf{Steps}} & \textbf{{Struc.}} &\multicolumn{4}{c|}{\textbf{Background Preservation}} &\multicolumn{2}{c|}{\textbf{Text Alignment}} &\multicolumn{1}{c|}{\textbf{IQA}} &\multicolumn{1}{c|}{\textbf{Temp.}} &\textbf{FiVE} \\
        & &Dist.$_{\times 10^3}$$\downarrow$ &PSNR$\uparrow$ &LPIPS$_{\times 10^3}$$\downarrow$ &MSE$_{\times 10^4}$$\downarrow$ &SSIM$_{\times 10^2}$$\uparrow$ &CLIPS.$\uparrow$ &CLIPS.$_{\text{edit}}\uparrow$ &NIQE$\downarrow$ & C.$_{\times 10^2}$$\uparrow$ &Acc$\uparrow$ \\
        
        \midrule
        \textbf{Ours} (\textbf{SF}) & 5 & 9.38 & 27.20 & 50.92 & 29.51 & 87.02 & 26.87 & 21.47 & 4.27 & 91.06 & 49.40 \\
        \textbf{Ours} (\textbf{SF}) & 9 & 8.90 & 28.39 & 46.58 & 21.92 & 87.90 & 26.87 & 21.73 & 4.32 & 91.37 & 48.74 \\
        \rowcolor{myhigh!50}
        \textbf{Ours} (\textbf{SF}) & 15 &{{10.27}} & {{28.47}} & {{49.84}} & {{21.88}} & {{87.52}} & {26.95} & {21.73} &{4.38} &{{90.93}} &{51.94} \\
        
        \midrule
        \textbf{Ours} (\textbf{LL})$^\S$ & 5 & 15.33 & 25.36 & 70.16 & 43.48 & 83.88 & 27.33 & 22.13 & 4.52 & 89.54 & 60.43 \\
        \rowcolor{myhigh!50}
        \textbf{Ours} (\textbf{LL})$^\S$ & 15 &{15.69} & {25.74} & {67.04} & {40.70} & {84.29} & {{27.22}} & {{22.11}} &{4.61} &{89.12} &{{61.19}} \\
        
        \bottomrule
    \end{tabular}}
    \label{tab:step}
\end{table*}

We further evaluation fewer-step editing results on FiVE-Bench, as shown in Tab. \ref{tab:step}.
Notably, even with reduced sampling steps, StreamEdit's overall performance on FiVE-Bench remains largely unchanged, maintaining comparable levels of both background retention and editing quality.
It fully demonstrates that the proposed generative paradigm stably leverages the outstanding properties of distilled streaming video generation models, resulting in overall performance that is insensitive to the number of sampling steps and facilitates fast, high-quality editing.
Meanwhile, increasing the number of sampling steps somehow slightly improves both background preservation and editing effectiveness, consistent with video generation models demonstrating better output quality as the number of sampling steps increases.
This further demonstrates that the proposed generative editing paradigm can better benefit from the development of generative models.

\begin{figure}[t]
\centering
\scriptsize
\begin{minipage}[t]{0.65\linewidth}
\centering
\captionof{table}{\textbf{Long-video editing} quantitative results.}
\vspace{1mm}
\resizebox{\linewidth}{!}{%
\begin{tabular}{l|ccc}
\toprule
\textbf{Methods}
& \textbf{Ewarp} $\times 10^{-3}$ $\downarrow$ 
& \textbf{CLIPS.} $\uparrow$ 
& \textbf{Temp. Cons.} $\uparrow$ \\
\midrule
StreamV2V & 2.53 & 28.45 & \underline{98.33} \\
AdFlow & \underline{1.65} & \underline{30.78} & 98.32 \\
\rowcolor{myhigh!50}
\textbf{Ours (LL)} & \textbf{0.68} & \textbf{31.25} & \textbf{99.23} \\
\bottomrule
\end{tabular}%
}
\label{tab:long_video_metric}
\end{minipage}
\hfill
\begin{minipage}[h]{0.33\linewidth}
\centering
\includegraphics[width=\linewidth]{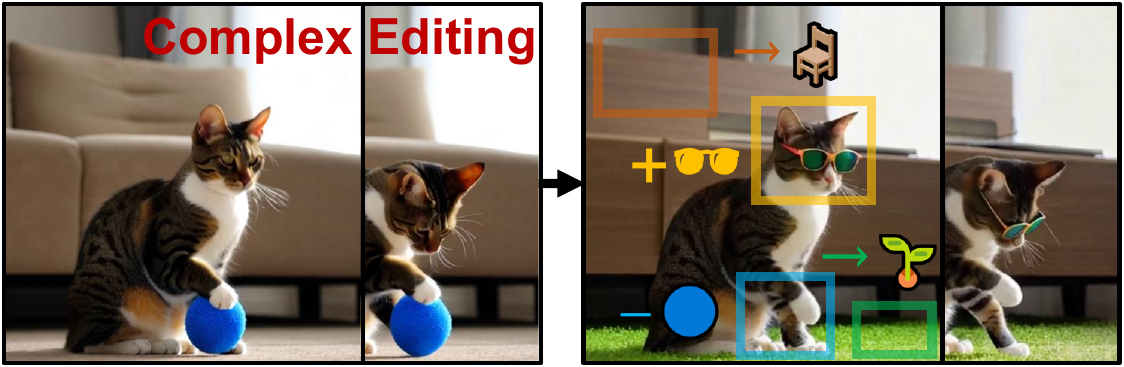}
\vspace{-0.5cm}
\captionof{figure}{\textbf{Chain editing} for complex objectives.}
\vspace{-3.25cm}
\label{fig:long_multi}
\end{minipage}
\end{figure}

\subsection{Long-Video Editing Quantitative Comparison}

In addition to user studies, we refer to existing approaches (\eg, Senorita \cite{zi2025senorita}) to compare Ewarp, CLIPScore, and Temporal Consistency for a total of over 10k frames on long-video editing tasks.
The results are shown in Tab. \ref{tab:long_video_metric}.
This result is aligned with user studies, further demonstrating that StreamEdit can better balance editing effectiveness and video quality compared to existing long-video editing methods.

\subsection{Chain Editing for Complex Editing Objectives}

When the goal of video editing is no longer a single concept, the editing objectives are often multiple and complex.
In this situation, thanks to StreamEdit's editing accuracy and fast sampling speed, we can achieve this goal through chain editing.
Fig. \ref{fig:long_multi} indicates a complex editing case consists of removing the toy ball, adding colorful sunglasses to the cat, changing the carpet to a lawn, and replacing the sofa with a desk.
Because StreamEdit has superior background preservation capabilities, there won't be significant changes to irrelevant regions after each edit. 
So we can gradually progress through each editing subtask, and then break down the complex editing objectives into multiple subtasks to achieve our goal.

\section{Analyses of Existing Training-Free Editing Paradigms}
\label{sec:appd_analyses}

\begin{table*}[t]
    \centering
    \caption{\textbf{Comparison of video editing on Self Forcing model}. 
    We reimplement FlowEdit and UniEdit-Flow on Self Forcing and evaluate them on FiVE-Bench.}
    \resizebox{0.99\textwidth}{!}{
    \begin{tabular}{l|c|cccc|cc|c|c|c}
        \toprule
        \multirow{2}{*}{\textbf{Methods}} & \textbf{{Struc.}} &\multicolumn{4}{c|}{\textbf{Background Preservation}} &\multicolumn{2}{c|}{\textbf{Text Alignment}} &\multicolumn{1}{c|}{\textbf{IQA}} &\multicolumn{1}{c|}{\textbf{Temp.}} &\textbf{FiVE} \\
        &Dist.$_{\times 10^3}$$\downarrow$ &PSNR$\uparrow$ &LPIPS$_{\times 10^3}$$\downarrow$ &MSE$_{\times 10^4}$$\downarrow$ &SSIM$_{\times 10^2}$$\uparrow$ &CLIPS.$\uparrow$ &CLIPS.$_{\text{edit}}\uparrow$ &NIQE$\downarrow$ & C.$_{\times 10^2}$$\uparrow$ &Acc$\uparrow$ \\
        
        \midrule
        FlowEdit (SF) & 21.58 & 22.04 & 143.15 & 76.42 & 73.45 & \underline{26.76} & \underline{20.89} & \textbf{4.23} & 86.45 & \underline{50.52} \\
        UniEdit-Flow (SF) & \underline{18.25} & \underline{26.84} & \underline{86.96} & \underline{26.69} & \underline{83.90} & 26.37 & 20.77 & \underline{4.30} & \underline{90.81} & 46.36 \\
        
        \midrule
        \rowcolor{myhigh!50}
        \textbf{Ours} (\textbf{SF}) &{\textbf{10.27}} & {\textbf{28.47}} & {\textbf{49.84}} & {\textbf{21.88}} & {\textbf{87.52}} & \textbf{26.95} & \textbf{21.73} &{4.38} &{\textbf{90.93}} &\textbf{51.94} \\
        
        \bottomrule
    \end{tabular}}
    \label{tab:paradigm_analyses}
\end{table*}

To validate the flaws inherent in existing paradigms for model development as discussed in this paper, we evaluate the performance of typical editing paradigms (inversion-based: UniEdit-Flow \cite{jiao2025uniedit}, inversion-free: FlowEdit \cite{kulikov2024flowedit}) using streaming video generation model (Self Forcing) on FiVE-Bench, as shown in Tab. \ref{tab:paradigm_analyses}.
Without altering the fact that these existing methods still require a relatively large number of sampling steps, the results they yield are inferior in most various respects to the proposed StreamEdit which is based on the introduced generative paradigm.
This experimentally corroborates our core motivation of proposing generation-style editing paradigm.

In principle, inversion is deterministic, and its accuracy depends on the number of sampling steps \cite{song2020denoising, ju2023direct, mokady2023null, jiao2025uniedit}. 
Therefore, even if models can generate outputs in a few steps through stochastic few-step sampling, inversion-based editing forces them to revert to deterministic sampling for video editing.
This prevents demonstrating that model development inherently improves the accuracy of deterministic inversion, thereby failing to ensure that editing can leverage the advancements of generative models.
Meanwhile, inversion-free editing essentially involves iteratively obtaining the mean trajectory toward the editing target from a set of random auxiliary trajectories \cite{infedit, kulikov2024flowedit, kim2025flowalign}.
Consequently, even when models can generate videos in just a few steps, these methods still require multiple iterations to obtain stable average trajectories.

Further modification to model architectures that better suit video editing tasks may help existing paradigms perform more effectively. 
The self-attention bridge proposed in this paper also represents an attempt to modify model architectures for video editing tasks. 
We hope our approach will bring both inspirations of extensions on architecture design to existing methods and improvements to editing paradigms.

\section{Details of Streaming Video Editing Comparison}
\label{sec:appd_detail_streaming}

\begin{figure*}[!t]
    \centering
    \includegraphics[width=0.99\linewidth]{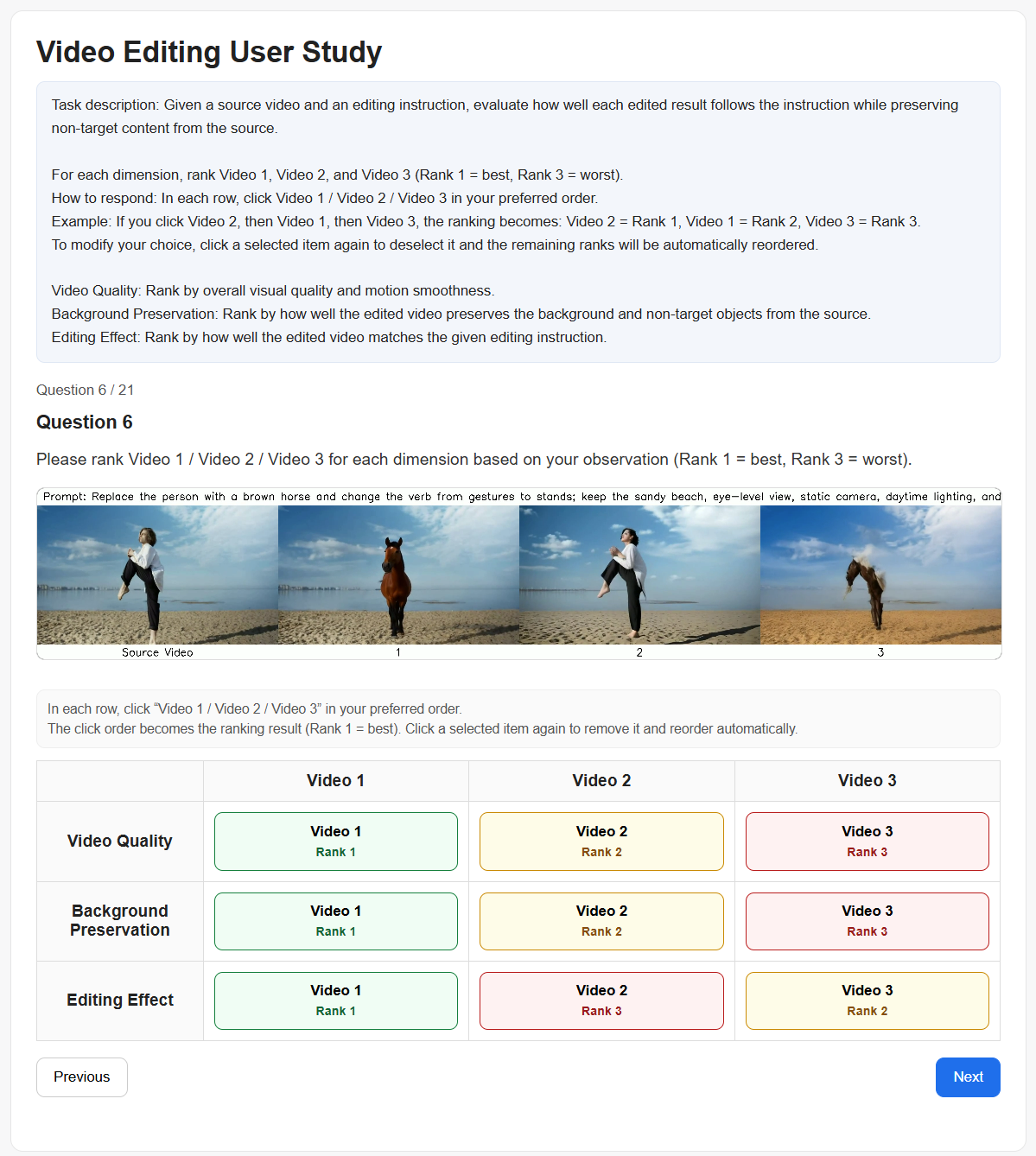}
    \caption{
    \textbf{User study instruction screenshots}.
    We shuffle the order of video samples and editing methods, then ask users to rank video quality, background preservation, and editing effect separately with reference to the source video and editing objective.
    }
    \label{fig:uspage}
\end{figure*}

\subsection{Implementation Details}

We prepare source prompts, target prompts, source trigger words, and target trigger words for the collected long videos (more than 470 frames), and perform 5-step video editing using the text-driven StreamEdit based on LongLive.
For long video editing baselines (StreamV2V \cite{liang2024looking}, AdaFlow \cite{zhang2025adaflow}), we employ their publicly available official implementations and maintain the hyper-parameter settings used in their published reports.

\subsection{User Study Details}

As shown in Fig. \ref{fig:uspage}, we build a webpage for the user study, featuring user instructions, video samples, and multiple-choice questions.
We conceal method names and randomly shuffle the order of video samples and methods for fairness.
The questions involve ranking the effectiveness of three aspects, measuring the relative performance of each method from perspectives of video quality, background preservation, and editing effect.

\section{Additional Qualitative Results}
\label{sec:appd_add_qualitative}

\subsection{General Video Editing}

\begin{figure*}[!t]
    \centering
    \includegraphics[width=0.99\linewidth]{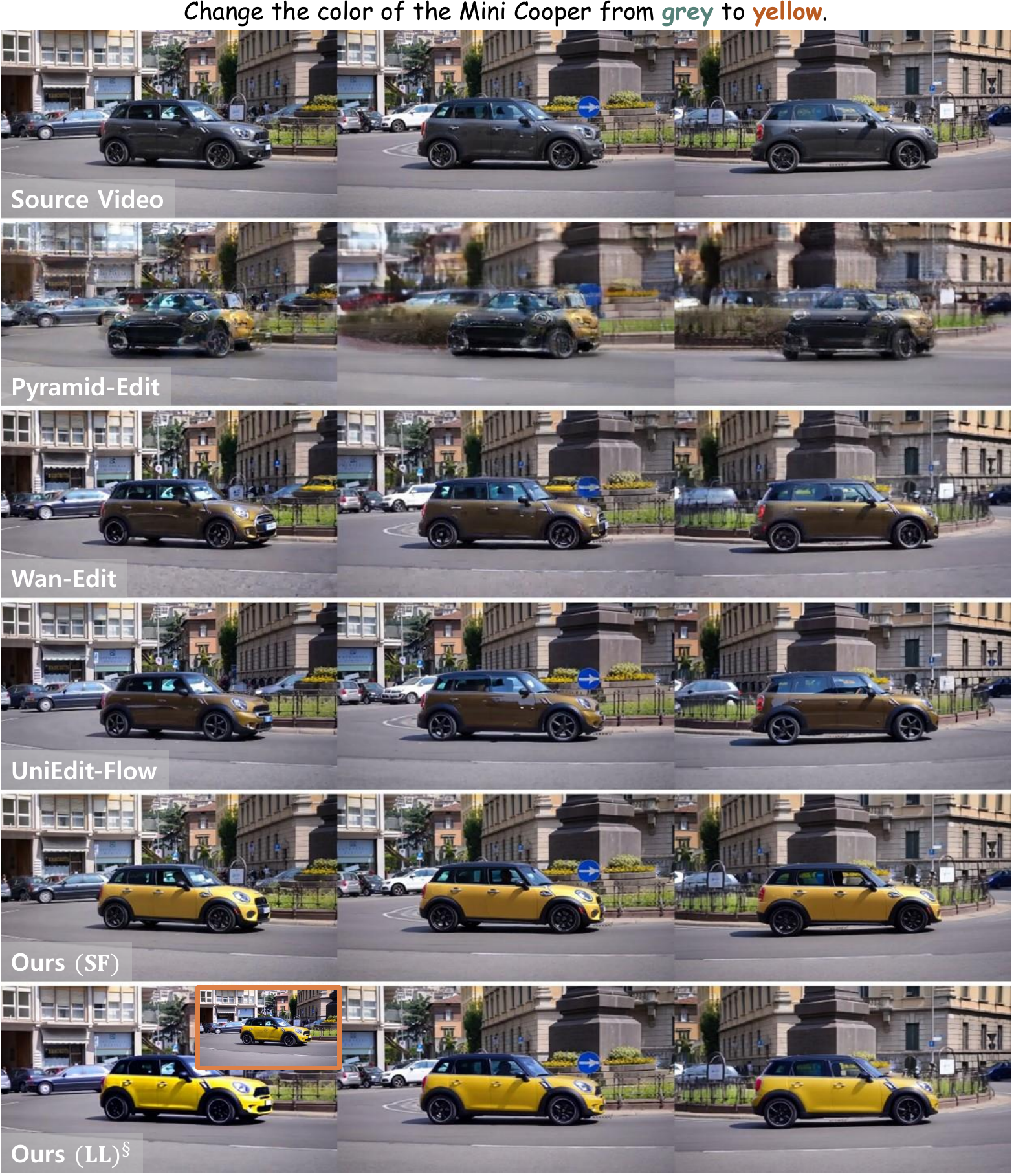}
    \caption{\textbf{Additional qualitative comparisons} of methods based on flow matching video generation models on the color editing case.
    Data-to-data editing starting from source videos may not be effective enough for significant color editing, but StreamEdit can successfully change the color.}
    \label{fig:appd_vis_1}
\end{figure*}

\begin{figure*}[!t]
    \centering
    \includegraphics[width=0.99\linewidth]{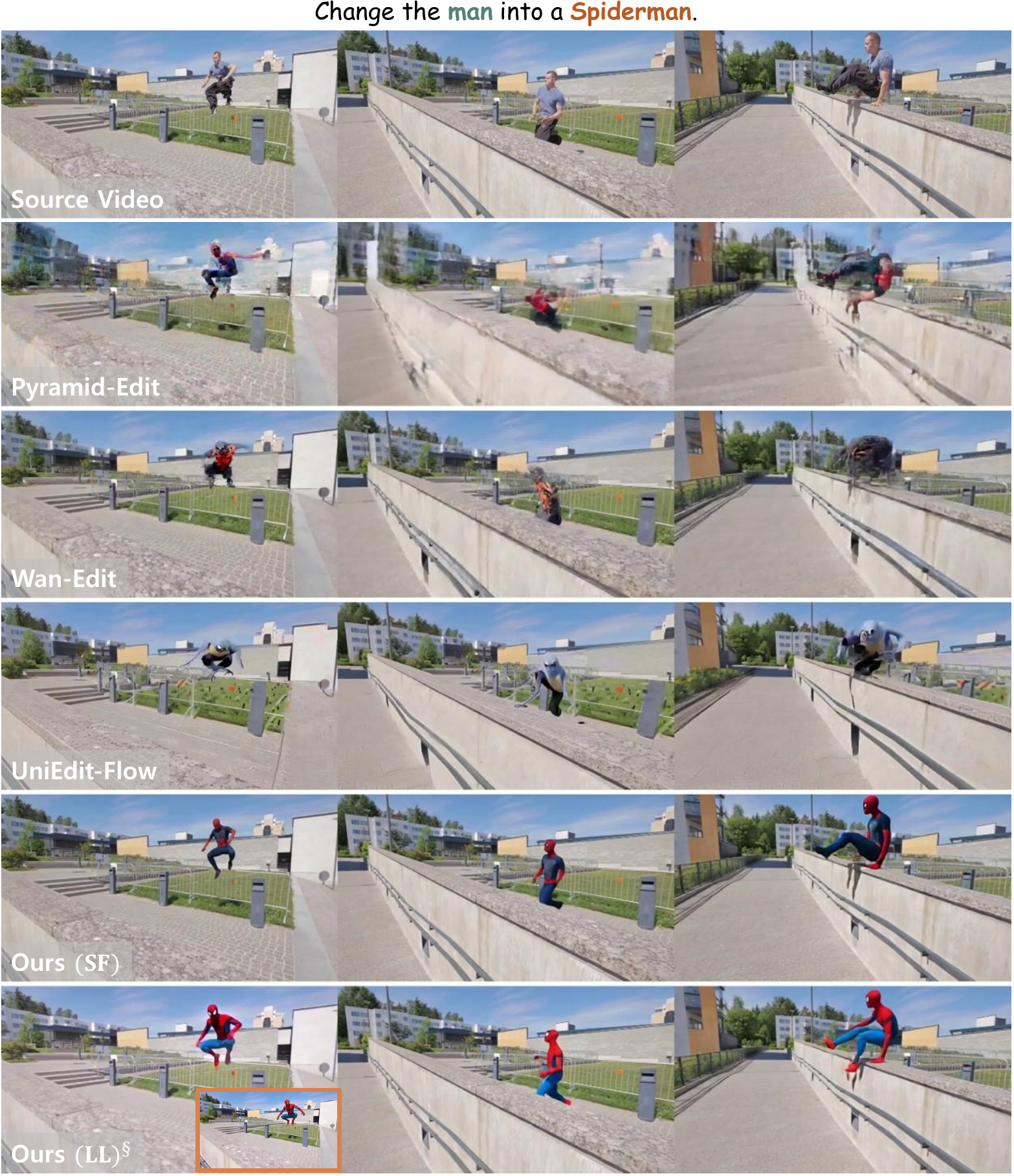}
    \caption{\textbf{Additional qualitative comparisons} of methods based on flow matching video generation models on the object modification editing case.
    StreamEdit better understands the editing objective and maintain high-quality editing results compared with existing methods.}
    \label{fig:appd_vis_2}
\end{figure*}

\begin{figure*}[!t]
    \centering
    \includegraphics[width=0.99\linewidth]{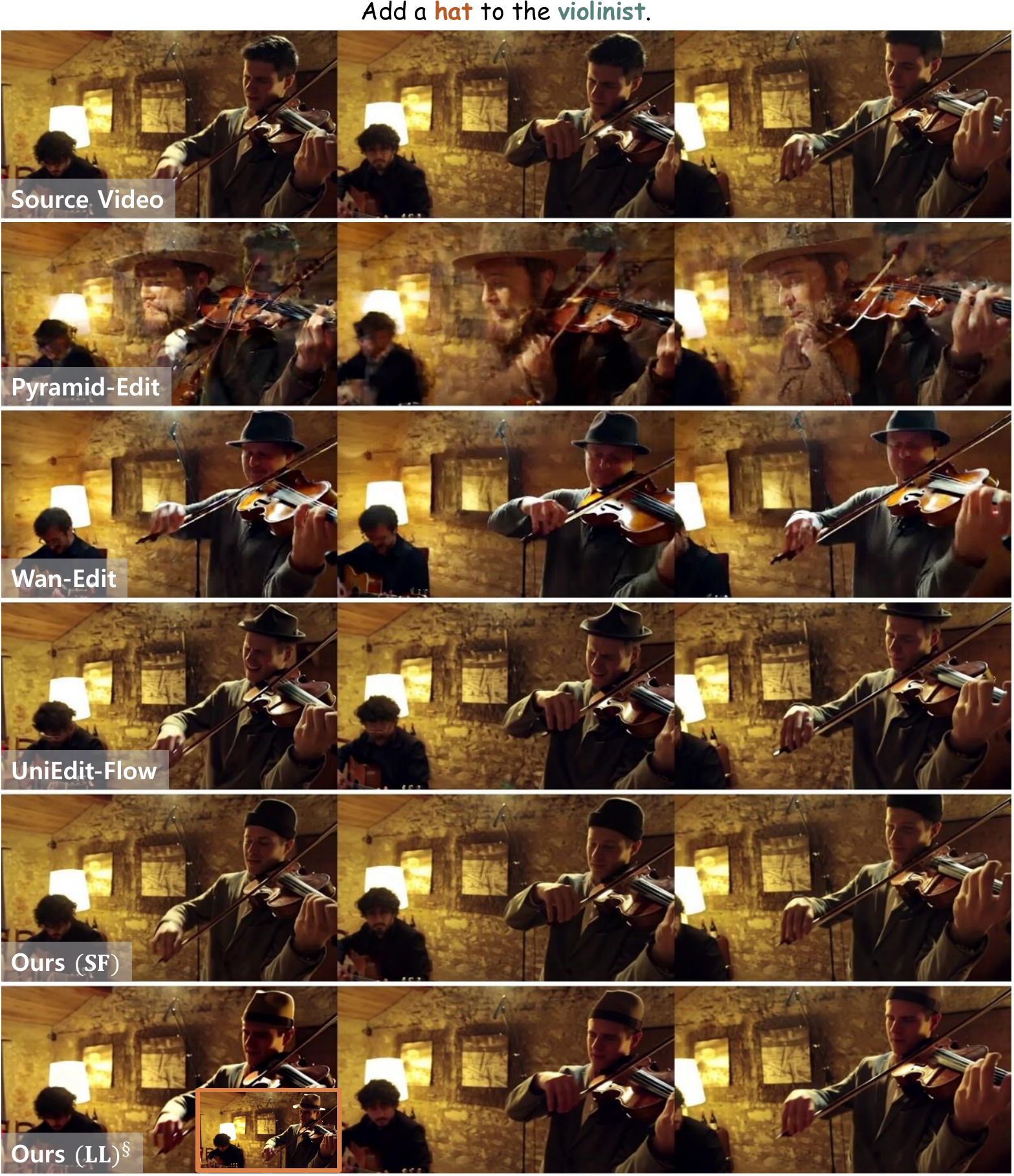}
    \caption{\textbf{Additional qualitative comparisons} of methods based on flow matching video generation models on the object addition editing case.
    Existing methods have difficulties to maintain consistency in background and identities, while StreamEdit ensures the background remains almost unchanged while successfully editing.}
    \label{fig:appd_vis_3}
\end{figure*}

\begin{figure*}[!t]
    \centering
    \includegraphics[width=0.99\linewidth]{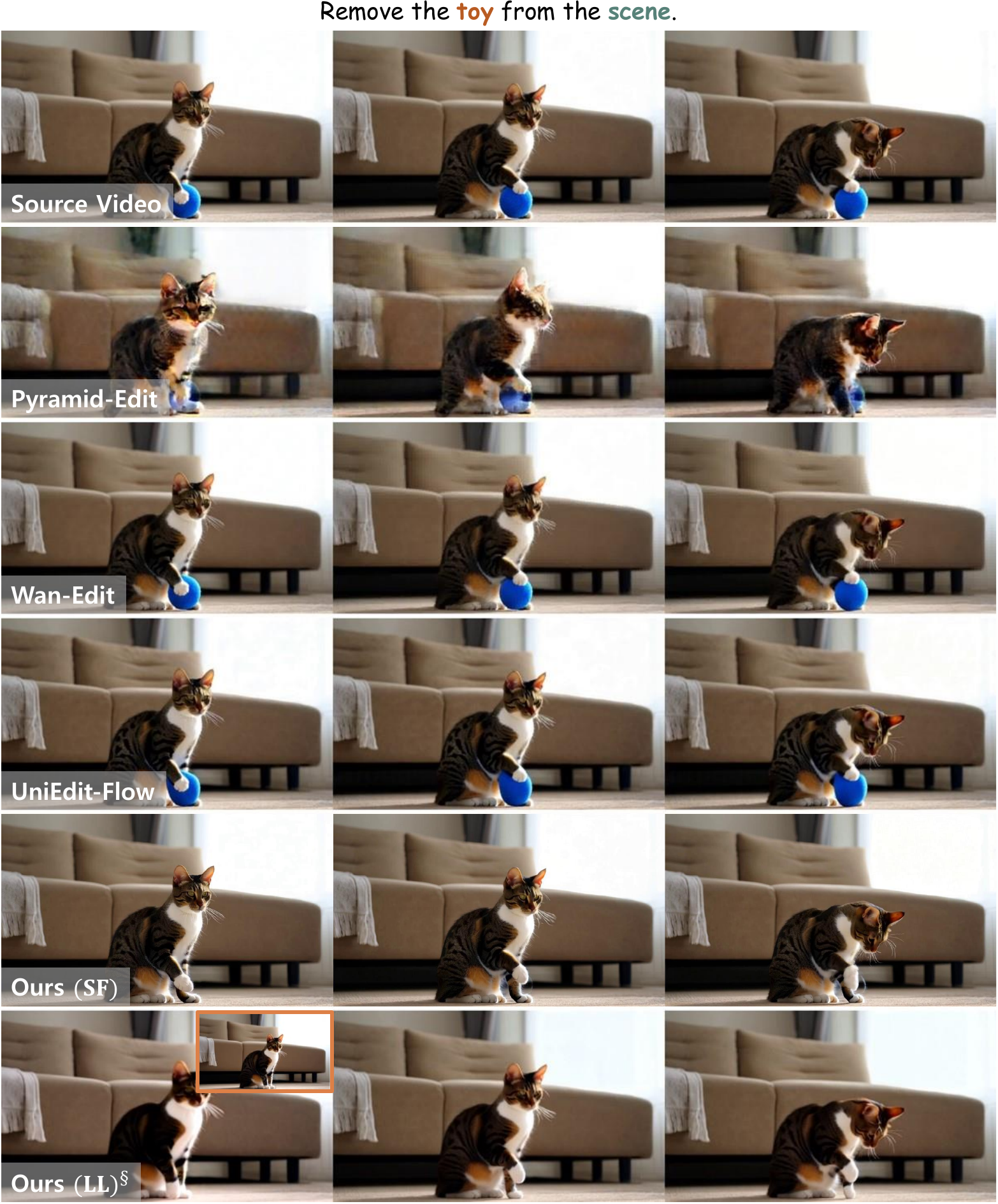}
    \caption{\textbf{Additional qualitative comparisons} of methods based on flow matching video generation models on the object removal editing case.
    Existing data-to-data methods cannot accurately eliminate the target object due to the target object to be eliminated being too conspicuous, while StreamEdit can perform editing accurately.}
    \label{fig:appd_vis_4}
\end{figure*}

We further provide qualitative comparisons of video color editing (Fig. \ref{fig:appd_vis_1}), object modification (Fig. \ref{fig:appd_vis_2}), object addition (Fig. \ref{fig:appd_vis_3}), and object removal (Fig. \ref{fig:appd_vis_4}).
Here we mainly compare the methods based on state-of-the-art flow matching video generation models \cite{li2025five, kulikov2024flowedit, jiao2025uniedit}.
These results further demonstrate that the proposed StreamEdit can successfully edit and preserve precise backgrounds on various editing tasks, demonstrating significant superiority over existing state-of-the-art methods.

\subsection{Text-Driven Streaming Editing for Long Videos}

\begin{figure*}[!t]
    \centering
    \includegraphics[width=0.99\linewidth]{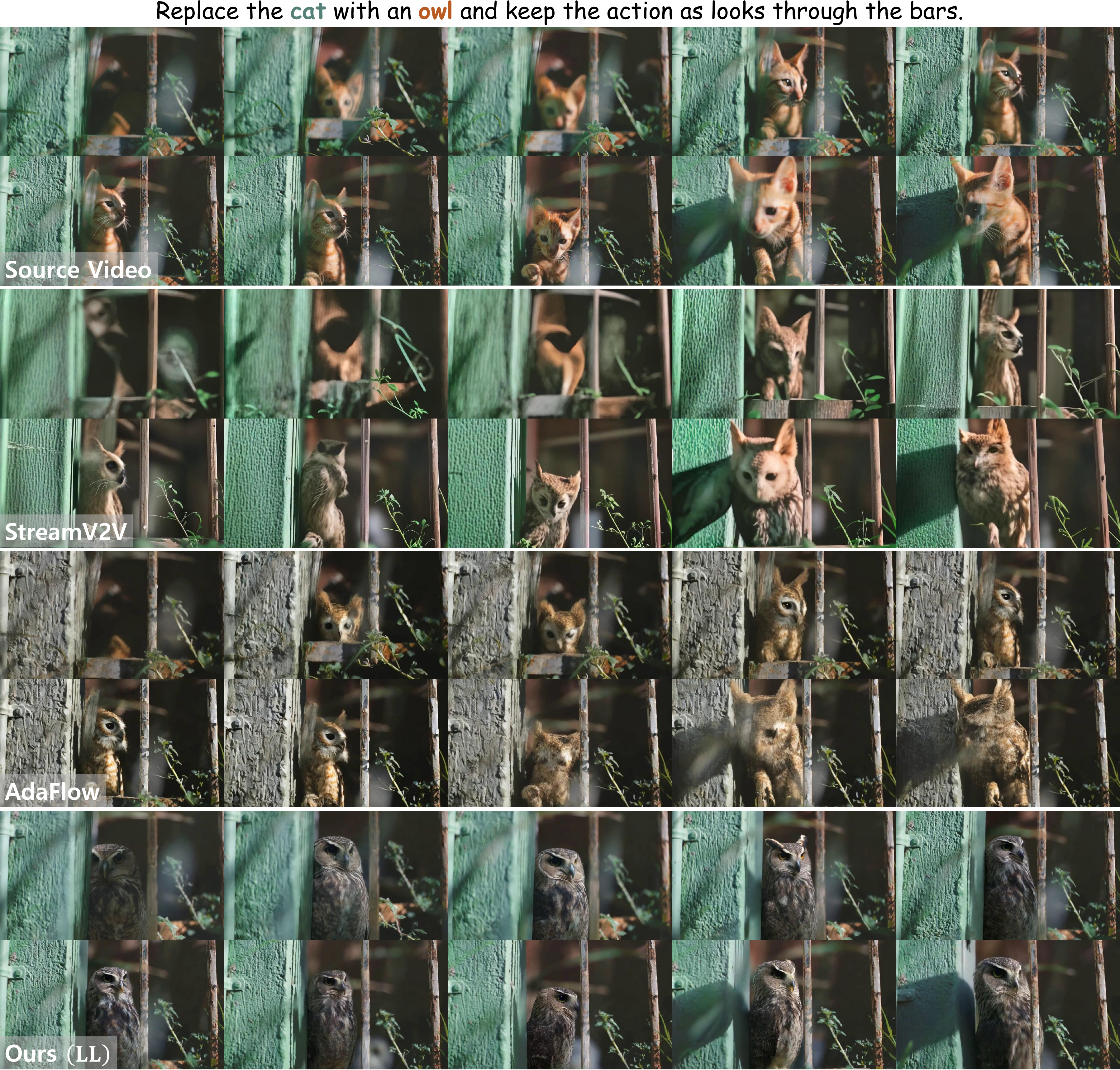}
    \caption{\textbf{Additional qualitative comparisons} of streaming editing for long videos.
    StreamEdit accurately and consistently achieves the editing objective while maintaining nearly lossless backgrounds, performing more effective than previous approaches.}
    \label{fig:appd_long_1}
\end{figure*}

\begin{figure*}[!t]
    \centering
    \includegraphics[width=0.99\linewidth]{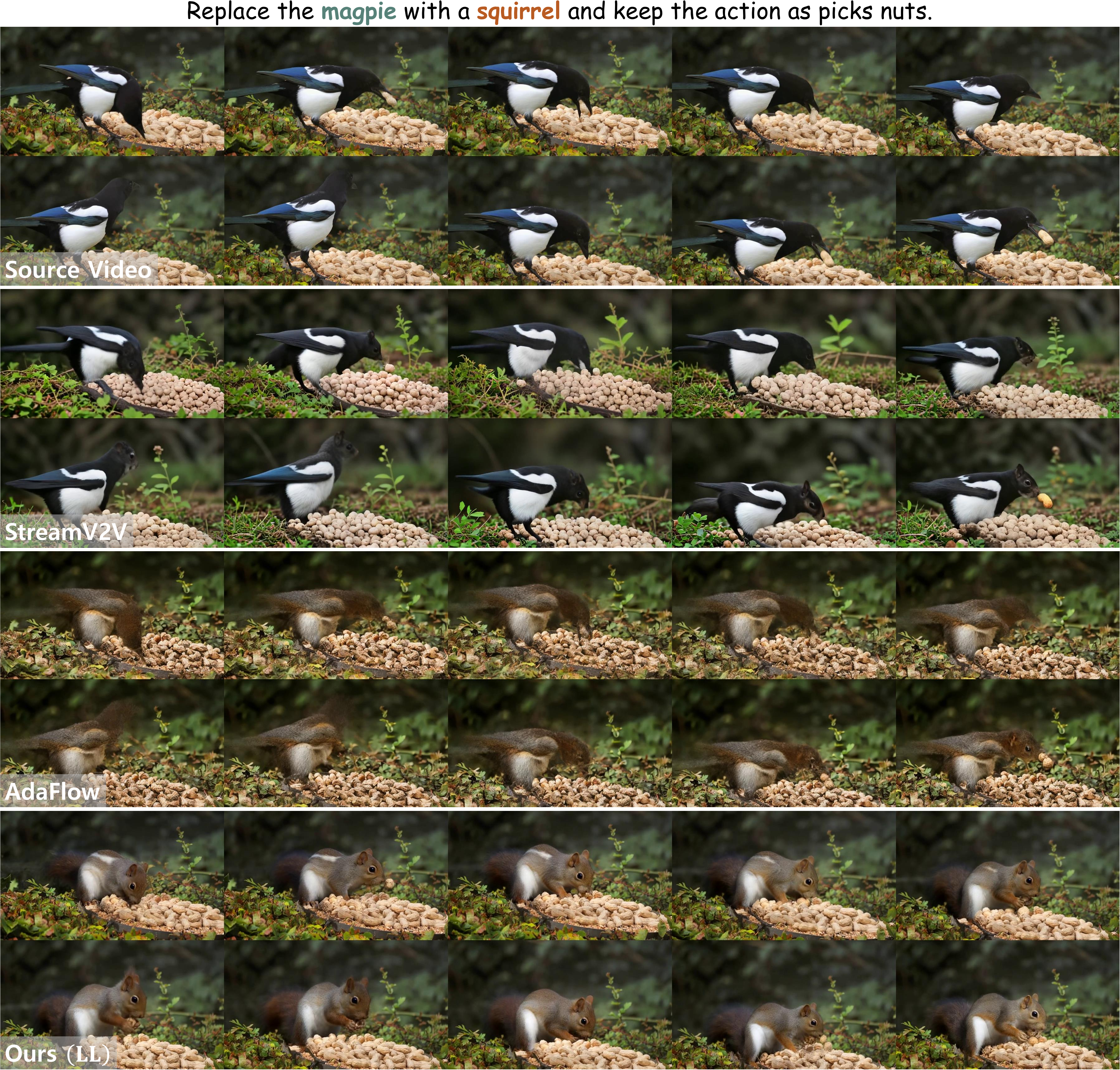}
    \caption{\textbf{Additional qualitative comparisons} of streaming editing for long videos.
    Benefiting from the generative video editing paradigm, StreamEdit enables the editing target to perform natural and plausible actions even when structural differences exist between the editing objective and the source object.}
    \label{fig:appd_long_2}
\end{figure*}

Fig. \ref{fig:appd_long_1} and \ref{fig:appd_long_2} provide more qualitative comparisons of streaming video editing tasks, showcasing that StreamEdit performs consistent and successful editing on long videos even when editing cases are challenging for existing long-video editing methods.

\section{Limitation and Future Work}
\label{sec:appd_lim}

\begin{figure*}[!t]
    \centering
    \includegraphics[width=0.99\linewidth]{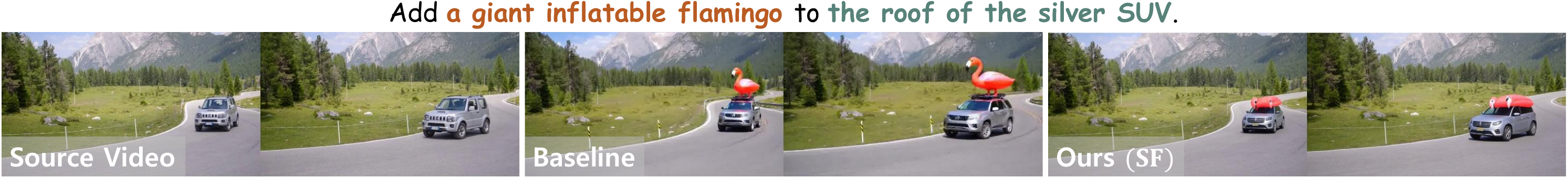}
    \caption{\textbf{Failure case} of the proposed StreamEdit.
    When facing large-scale additions, StreamEdit's focus on precision editing makes it less effective at applying edits to sufficiently large regions, resulting in less visually satisfying added objects.}
    \label{fig:appd_fail}
\end{figure*}

As shown in Fig. \ref{fig:appd_fail}, we provide a typical failure case of StreamEdit.
Since StreamEdit prioritizes precise video editing, it may struggle to deliver sufficiently noticeable editing effects when dealing with large-region additions, potentially resulting in less visually satisfactory results in such scenarios.
Moreover, for fast-moving objects or multi-shot videos, the editing results sometimes may still contain artifacts or changes, largely due to inheriting the performance shortcomings of existing autoregressive models. 
The potential future solution is to use more advanced streaming video generation models or introduce architectures with consistency or memory modules (such as MemFlow \cite{ji2025memflow}).
Furthermore, we will continue to explore the design of attention mechanism manipulation, striving to make the generative editing paradigm more robust and effective.

Additionally, the proposed few-step streaming editing method StreamEdit can be regarded as a conditional video generation model. 
Therefore, a potential possibility is to leverage it as a teacher model to guide training-based video editing approaches.
We consider it an intriguing future work.

\end{document}